\documentclass[journal]{IEEEtran}

\usepackage{cite}
\usepackage{amsmath,amssymb,amsfonts}
\usepackage{algorithmic}
\usepackage{graphicx}
\usepackage{textcomp}
\def\BibTeX{{\rm B\kern-.05em{\sc i\kern-.025em b}\kern-.08em
    T\kern-.1667em\lower.7ex\hbox{E}\kern-.125emX}}

\usepackage{caption}
\usepackage{subcaption}

\usepackage{gensymb}
\usepackage{graphicx}
\usepackage{subcaption}
\usepackage{cite}
\usepackage{color, colortbl}
\usepackage{url}
\usepackage{bm}
\usepackage{array}

\usepackage{tikz}
\usepackage{pgfplots}
\usepgfplotslibrary{groupplots}
\usepgfplotslibrary{external}
\usetikzlibrary{pgfplots.groupplots}
\usetikzlibrary{patterns} 
\usepackage{placeins}
\usepackage{multirow}

\usepackage{times} 

\usepackage{amsbsy}  
\usepackage{colortbl}

\definecolor{lightgray}{rgb}{0.9,0.9,0.9}
\definecolor{mumred}{rgb}{0.863,0.129,0.302}
\definecolor{mumgreen}{rgb}{0,0.549,0}   
\definecolor{mumblue}{rgb}{0,0.392,0.871} 
\definecolor{mumpurple}{rgb}{0.4,0.0,0.4} 
\definecolor{mumorange}{rgb}{1,0.4,0} 
\definecolor{mumteal}{rgb}{0.216,0.784,0.671} 
\definecolor{lightgrayB}{gray}{0.95} 
\DeclareMathOperator*{\argmin}{arg\,min}

\usepackage{hyperref}
\usepackage{units}

\begin{document}

\title{Experimental Study on Reinforcement Learning-based Control of an Acrobot}
\author{{Leo Dostal$^a$}, {Alexej Bespalko$^a$, and Daniel A. Duecker$^a$} \\
\vspace{0.2cm}$^a$\textit{Hamburg University of Technology, Institute of Mechanics and Ocean Engineering, \\
Eissendorfer Strasse 42, 21073 Hamburg, Germany}
}
 




\maketitle

\begin{abstract}
We present computational and experimental results on how an artificial intelligence (AI) learns to control an Acrobot using reinforcement learning (RL). 
For this, the experimental setup is designed as an embedded system, which is of interest for robotics and energy harvesting applications. 
Specifically, we study the control of angular velocity of the Acrobot, as well as control of its total energy, which is the sum of the kinetic and the potential energy.
\textcolor{black}{We use an RL approach to these control problems, as it circumvents the user to derive the Acrobot's complicated equations of motion.}
By this means the RL algorithm is designed to drive the angular velocity or the energy of the first pendulum of the Acrobot towards a desired value. 
With this, libration or full rotation of the unactuated pendulum of the Acrobot is achieved. 
Moreover, investigations of the Acrobot control are carried out, which lead to insights about the influence of the state space discretization, the episode length, the action space or the mass of the driven pendulum on the RL control. 
By further numerous simulations and experiments the effects of parameter variations are evaluated. \textcolor{black}{Finally, we compare the achieved performance of the RL controller to results of a sliding mode controller. Both controllers are able to stabilize the Acrobot at a desired energy with comparable performance. Thereby, it turns out that the RL controller has some advantages compared to the sliding mode controller.}
\end{abstract}




\section{INTRODUCTION}
An essential component of conventional control technology is the derivation of a mathematical model in order to describe the real behavior of the controlled system as well as possible.
The design of a controller usually goes hand in hand with complex non-linear differential equations.
In practice, the involved parameters can only be determined with limited precision and usually with a lot of effort. 
Moreover, these parameters apply only under certain conditions. 
Especially the control of nonlinear underactuated systems is not easily achieved by control algorithms, if the dynamical behavior is complicated and difficult to model. 
For such problems finding an accurate mechanical model for a control task is often very time consuming or er even not feasible. 
To avoid these circumstances, alternative methods from the field of machine learning may be worthwhile. 
Recent advances enable innovative and economical control of complex systems. 
Intelligent processes that were considered inefficient many years ago are now gaining new research attention and form a basis for new expanding work. 
This trend is strongly featured by recent advances in computing power. 
Hereby, especially the trend of miniaturization allows energy-efficient computing units which allow at-the-edge-computing rather than transferring data towards a server.
In this sense we aim to study whether reinforcement learning (RL) based control without prior modelling of the underactuated system is possible in this context. 
As a particular underactuated system we use the Acrobot \cite{spong1995swing} in our study due to the complexity of its nonlinear behavior, which makes it possible to adapt our results to various other systems in various fields, such as mechanical, MEM, robotic, and other systems. Moreover, energy generation from ocean gravity waves through pendulum excitation has become an emerging topic of interest, since gravity waves can be used to excite the pivot of a pendulum in order to induce its oscillation or rotation, which can be used for energy harvesting~\cite{vaziri2014experimental,alevras2015experimental,dostal:2017b}. 
Energy can be harvested from the system using electromagnetic induction. 
Pendulum motion control laws can be deployed in order to maximize the harvested energy. Hereby, the pendulum motion controller aims to find a policy which effectively drives the pendulum close to its eigenfrequency which maximizes the harvested energy. 
These are referred to as 2:1 parametric resonance -- the mean period of the excitation is twice the period of the response, cf. \cite{dostal:2017b}. This system can be controlled, if a second smaller pendulum is attached to the energy harvesting pendulum. The resulting system can then be identified as an Acrobot.
For the Acrobot, various control strategies have been used for its control \cite{brown1997intelligent,awrejcewicz2012experiment,lee2015robust}.
However, prior work usually focuses on the task of stabilizing the Acrobot at its inverted position, referred to as balancing control.
%
%
%
%
Instead of this we aim to control the system's energy level in the presented work \textit{without} assuming any prior knowledge on the system itself. Thereby the control algorithms are implemented on a low cost embedded platform, which has very limited computational power. This is important for specific applications in robotics or energy harvesting. This restriction limits the possibilities of using complex control algorithms such as model predictive control.  
For the control of the Acrobot without prior knowledge of its model parameters and using a low power embedded platform with very limited computational power we therefore study a control strategy based on RL.
Recently, RL control was adapted to the experimental setup from \cite{dostal2019theoretical} in order to positively influence its energy state \cite{cyr2019towards}, by means of energy control of an Acrobot. \textcolor{black}{However, in the preliminary study \cite{cyr2019towards}, we have only used one setup for the simulation and one setup for the experiment. Thereby, only one desired energy level was tested in simulation and experiment. In that study, it was not clear how the RL controller behaves for other setups or different desired energy levels. Moreover, the achieved performance was not evaluated in comparison to a conventional control method.
Therefore, further investigations with the Acrobot are carried out in this study. 
This should lead to new insights about the influence of the state space discretization, the episode length, the action space or the mass of the driven pendulum on the RL algorithm. By further numerous simulations and experiments the effects of parameter variations are evaluated. Afterwards, the combination of interesting results is presented.
In this sense, we study whether RL based control is feasible for energy-level control of the Acrobot. Thereafter, we compare the achieved performance to a sliding mode controller in section~\ref{sec:sliding_mode_control}, since the sliding mode controller is a conventional control method for nonlinear dynamical systems. Then we conclude our findings in section~\ref{sec:conclusions}.}

\section{PROBLEM STATEMENT}\label{sec:problemstatement}
\textcolor{black}{The Acrobot is a challenging control problem in Control Theory. 
It is an underactuated non-linear system, which consists of two links and two joints. Thereby, only the joint between the two links can be actuated by applying a torque. The system is mounted at the other joint without a possibility of actuation. The name Acrobot is a fusion of the words acrobat and robot, since the system looks like an acrobat swinging on a high bar.}
\textcolor{black}{This robot has been also studied in the context of machine learning} \cite{DAMOTTASALLESBARRETO2008454,brown1997intelligent,boone1997efficient,boone1997minimum}.

\begin{figure}
    \footnotesize
    \begin{center}
        \def\svgwidth{0.95\columnwidth} 
        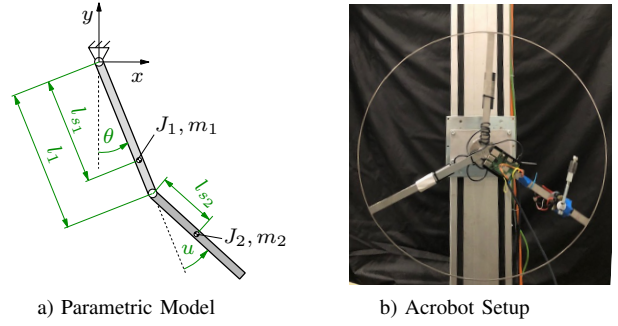
        \setlength{\belowcaptionskip}{-10pt}
        \caption{Panel (a): Model of the Acrobot. Panel (b): Physical setup with RaspberryPI\,3B+, IMU, and USB-powerbanks mounted on the flywheel. A video of the experimental setup can be found under: \url{https://youtu.be/kca-Gpdmp3I}\\
        $ $}
        \label{fig:doublependulum}
    \end{center}
\end{figure}

\textcolor{black}{In the right panel of Fig.~\ref{fig:doublependulum} the Acrobot is shown, on which the experiments in this work are performed. 
 It consists of a flywheel with the mass $ M $ and moment of inertia $ J_{SR} $, which is mounted at the pivot in its center in such a way that it can freely rotate. At a distance $ l_{1} $ from the center a servo motor including an additional weight with the mass $ m_M $ is attached, which drives a thin rod with the length $ l_{2} $ and mass $ m_S $. A point mass $ m_p $ is also located on the tip of the rod. The RL algorithm is executed on a Raspberry Pi 3 B+ with the mass $ m_{pi} $, which has the distance $ l_{pi} $ from the pivot. Two batteries are fixed on the other two ribs of the flywheel for the power supply. These have the same distance $ l_A $ from the suspension point of the flywheel and are assumed as point masses $ m_A $. The total center of gravity of the servo motor, the batteries and the computer is located at a distance $ l_{c_1} $ from the center of rotation, so that this setup is equivalent to a physical pendulum with the total mass $ m_1 =  2m_A+m_M+m_{pi} $. Thereby, $ l_{c_1} $ is given by
 \begin{equation}
 	l_{c_1} = \dfrac{l_1m_M + l_{pi}m_{pi}- 2l_Am_A\sin(30°)}{m_1}.
 \end{equation}
 All of the above-mentioned quantities have been determined and are given in Tab~\ref{tab:params}. A model of this arrangement is shown in panel (a) of Fig.~\ref{fig:doublependulum}.} 
%

\textcolor{black}{After evaluation of the corresponding Lagrange's equations of the second kind
and the assumption of linear damping, the equations of motion for the Acrobot with actuated second arm are given by}
\begin{equation}\label{eq:AcrobotEoM}
\begin{aligned}
&(J_1 + J_2 + m_2 ( l_1^2 +2 l_1 l_{c_2}\cos(u))) \ddot{\theta} = - m_2 l_{c_2} g \sin(\theta+u) \\
&- (m_1 l_{c_1} + m_2 l_1) g \sin(\theta) + m_2 l_1 l_{c_2} \sin(u) (\dot{u}^2+ 2 \dot{\theta}\dot{u})   \\
& -(J_2+ m_2 l_1 l_{c_2}\cos(u))\ddot{u}-d_1\dot{\theta},\\
&(J_2 + m_2l_1l_{c_2}\cos(u))\ddot{\theta} + J_2\ddot{u} \\
&+ m_2l_1l_{c_2}\sin(u)\dot{\theta}^2 + m_2gl_{c_2}\sin(\theta + u) + d_2\dot{u} = M_u.
\end{aligned}
\end{equation}
\textcolor{black}{with the angular position $\theta$ of the first pendulum, the mass $m_2= m_s+m_p$ and angular position $u$ of the second pendulum, damping coefficient $d_1$, and acceleration due to gravity $g$. An over-dot $\dot{()}$ denotes differentiation with respect to time $t$. }

\textcolor{black}{The goal in this study is to use energy level control of the Acrobot instead of other control tasks such as balancing control. In order to be able to apply the energy level control, we have chosen the total energy $H$ of the first pendulum as the target energy to be reached. This energy is the sum of the kinetic and potential energy of the first pendulum, which is given by the Hamiltonian}
\begin{equation}\label{eq:hamiltonian}
H=\frac{1}{2}J_1\dot{\theta}^2+2 m_1 l_{c_1} g \sin^2\left(\frac{\theta}{2}\right).
\end{equation}
\textcolor{black}{If the total energy of the full Acrobot system, which includes also the actuated second pendulum, would have been chosen as the target energy, then it would be possible to reach the target energy by just rotating the actuated second pendulum, which is simple and not of interest for us.}

\textcolor{black}{It is evident from Eq.~\eqref{eq:AcrobotEoM} that the resulting nonlinear dynamics of the Acrobot are not trivial and lead to a challenging control problem, which results in a significant effort for classical controller design. This is due to the fact that a  
transition at a separatrix between the two different dynamical regimes of libration and rotation of the first pendulum exist, combined with highly nonlinear behavior of the system, see for example}~\cite{dostal:2017b}. 
\textcolor{black}{In the present work we are interested to explore whether it is possible to circumvent the derivation of equations of motion for controller development even for such difficult control problems like the present Acrobot control. For this, we use the RL framework.}

\begin{table}
    \setlength{\belowcaptionskip}{-10pt}
    \caption{Parameters of the physical setup.}
    \label{tab:params}
    \begin{center}
        \begin{tabular}{|c|c|}
            \hline 
            \textbf{object} & \textbf{parameter} \\ 
            \hline 
            flywheel & $ J_{SR} = \unit[0{,}036]{kgm^2} $\\ 
            \hline 
            servo motor & $ m_M = \unit[0{,}158]{kg}, l_1 = \unit[0{,}26]{m} $\\ 
            \hline 
            battery cell & $ m_A = \unit[0{,}133]{kg}, l_A = \unit[0{,}15]{m} $\\ 
            \hline
            Raspberry Pi & $ m_{pi} = \unit[0{,}042]{kg}, l_{pi} = \unit[0{,}07]{m}$\\
            \hline 
            small pendulum: mass & $ m_p = \unit[0{,}038]{kg} $\\ 
            \hline 
            small pendulum: rod & $ m_r = \unit[0{,}042]{kg}, l_2 = \unit[0{,}1]{m} $  \\ 
            \hline 
        \end{tabular}
    \end{center}
\end{table}

\section{REINFORCEMENT LEARNING FRAMEWORK}\label{sec:reinforcementLearning}
In this section, we briefly recap the RL framework presented in \cite{brinkmann2018reinforcement} and its adaptation to pendulum energy control presented in our earlier work \cite{cyr2019towards}. 
Thereby, we concentrate on  the main steps of the original algorithm and highlight our key extensions. 
We refer the reader to \cite{sutton2018reinforcement, kaelbling1996reinforcement} for a detailed description of the underlying algorithm.

We observe that the underlying energy control problem can be written as a Markov decision process~(MDP).
It is defined by the discrete states~$S$, the actions~$A$, an energy-based rewards function $R(s,a)$, and a state transition matrix~$\boldsymbol{P}(s^\prime|s,a)$. 
This transition probability matrix represents the algorithm's belief on the expected successor state $s^\prime$ given the current state $s\in S$ and a selected action $a\in A$.
Therefore, the RL-algorithm chooses $a$ based on the reward function.

We define a value function~$v(s)$ which determines the value of the single state~$s$.
Moreover, the state-action value function~$q\left(a,s\right)$ describes the value of the corresponding state-action pair. 
Note that our goal is to find the optimal state-action value function~$q^*\left(s,a\right)$.
This, can be determined via iterating over the Bellmann equation. 
Hence, the optimal policy~$\pi(s)$ can be obtained from
\begin{align}
\pi(s) = \argmin_a q^* \left(s,a\right).
\end{align}

Initially, the model is unknown to the RL-algorithm.
Thus, a fundamental trade-off arises between gathering system information to find good actions (\textit{exploration}) and the selection of good actions which drive the system to the desired goal (\textit{exploitation)}.
This is done in an iterative manner as we update parameters and evaluate the policy at the end of each episode.

\textcolor{black}{The state-space $S$ is discretized into states consisting of intervals of the Acrobot's angular position $\theta$ and its angular velocity $\dot{\theta}$.} 

Note that at a first glance, a finer state discretization is likely to improve the algorithm's performance. 
However, this becomes challenging for two reasons: 
First, the algorithm requires additional efforts for system model exploration, as revisiting of states becomes less likely. 
Second, deployment on embedded hardware likely renders infeasible due to its very limited computational power data storage. 

\textcolor{black}{The action space $A$ of the Acrobot only consists of actions $a\in A$ for the lever angle $u\in\left[\pi/2, 3\pi/2\right]$ of the second pendulum, since the angle $\theta$ of the first pendulum can not be actuated.}


We design the rewards function  such that it punishes deviation between the first pendulum's energy level Eq.\,\eqref{eq:hamiltonian} and the desired energy level $H_d$ reading
\begin{align}\label{eq:rewardfunctionstandard}
R(s^\prime,a) = -(H-H_d)^2.
\end{align}
Hereby, we weigh larger deviations from the nominal value stronger, note the square in Eq.\,\eqref{eq:rewardfunctionstandard}. 
By solving for $ H (\theta, \dot{\theta}) $, the reward received can be converted back into the current energy level $H$.
If the terminal state~$s_T$ is reached, a high negative reward is assigned in order to penalize control leading out of the discretized region.
Consider the general case of the Acrobot's parameters, such as $I_1$, $l_1$, etc., being unknown. 
Hereby, it is appealing to use the generic scaled energy $\tilde{H}$, within the reward function,
\begin{equation}
\tilde{H}=\frac{1}{2}\dot{\theta}^2+c_\mathrm{exp}\,2\sin^2\left(\frac{\theta}{2}\right),\label{eq:scaledhamiltonian}
\end{equation}
where $\tilde{H}$ is proportional to the energy $H$. 
Note that,  we have to identify the coefficient $c_\mathrm{exp}$ experimentally.
This is described in the following.
First, the Acrobot is initiated at an arbitrary starting angle $\theta_s$ resulting in an oscillating but non-rotating motion. 
We measure the angle $\theta_\mathrm{meas}$ when the angular velocity $\dot{\theta}$ equals zero.
Second, we measure $\dot{\theta}_\mathrm{meas}$ at $\theta=0$.
The measured values are substituted into Eq.\,\eqref{eq:scaledhamiltonian}, leading to
\begin{equation}\label{eq:measuredhamiltonian1}
\tilde{H}(0,\dot{\theta}_\mathrm{meas})=\frac{1}{2}{\dot{\theta}_\mathrm{meas}}^2
\end{equation}
and
\begin{equation}\label{eq:measuredhamiltonian2}
\tilde{H}(\theta_\mathrm{meas},0)=c_\mathrm{exp}\,2\sin^2\left(\frac{\theta_\mathrm{meas}}{2}\right).
\end{equation}
Since the energy $\tilde{H}(\theta,\dot{\theta})$ is expected to be constant we get
\begin{equation}\label{eq:consthamiltonian}
\tilde{H}(0,\dot{\theta})=\tilde{H}(\theta,0).
\end{equation}
\textcolor{black}{Then we obtain the formula for the coefficient $c_\mathrm{exp}$ as}  
\begin{equation}\label{eq:expFactor}
c_\mathrm{exp}=\frac{\frac{1}{2}\dot{\theta}^2_\mathrm{meas}}{2\sin^2\left( \frac{1}{2} \theta_\mathrm{meas} \right)}
\end{equation}
by substituting Eq.\,\eqref{eq:measuredhamiltonian1} and \eqref{eq:measuredhamiltonian2} into Eq.\,\eqref{eq:consthamiltonian}.
We find it straight forward to determine the energy $\tilde{H}_{\theta_0}$ in a calibration step by starting at the angle $\theta_0$ and angular velocity $\dot{\theta}_0=0$, followed by measuring $\dot{\theta}_\mathrm{cal}$ at $\theta=0$. 
Hence,
\begin{equation}\label{eq:Calibration}
\tilde{H}_{\theta_0}=\frac{1}{2}{\dot{\theta}_\mathrm{cal}}^2.
\end{equation}

As discussed before, the RL-algorithm runs in episodes.
An episode is terminated if either the terminal state or the maximum episode length is reached.
At each episode's begin, the algorithm determines its current state $s$ by measuring the angle $\theta$ and the angular velocity $\dot{\theta}$. 

Note that the state space is discrete. 
Thus, applying an action $a$ may or may not change the system state.
The successor state $s^\prime$ is obtained as before from new measurements of $\theta$ and $\dot{\theta}$ respectively. 
This is conveniently done at the next episode's time step and we give the corresponding reward $R(s^\prime,a)$ to the algorithm and $s^\prime$ becomes $s$ of the new time step.

The selection of an action is performed based on the current state $s$, the transition probability matrix $\boldsymbol{P}(s^\prime|s,a)$ and the state reward matrix $\boldsymbol{R}(s,a)$. 
Hereby, an action $a_i$ is favored based on the expected reward $\boldsymbol{R}(s,a_i)$ given the current dynamics belief $\boldsymbol{P}(s^\prime|s,a)$. 
However, beside the direct next state action selection affects the gathering of potential future rewards given this state.
These potential rewards are captured by the state value $v(s)$. 
Anyhow, since we prefer rewards gained in the near future over long-term gains we discount future rewards by a depreciation factor $\gamma$. 
Revisiting the trade-off between exploration and exploitation, the action selection is altered with a probability $p_{\text{explore}}$.
This actively enforces an exploration of the surrounding state space even at later episodes. 
\textcolor{black}{Note that the algorithm has initially no prior knowledge on the system parameters. 
Hence, altering the selected action is important as it explores and expands the validity of the dynamics model $\boldsymbol{P}(s^\prime|s,a)$ within the whole state space.} 

Within a single episode all state transitions, actions and their corresponding rewards are logged. 
When the episode is terminated, the value function function $v(s)$, the state reward matrix $\boldsymbol{R}(s,a)$, and the state transition matrix $\boldsymbol{P}(s^\prime|s,a)$ are update accordingly to the logged state transitions. 
Finally, the next episode is initialized.

At the terminus of episode $k$, policy iteration is performed to compute an improved state value function $v_{k+1}(s)$ for the subsequent episode. 
For this, incorporating the improved dynamics model from the previous step improves the prediction accuracy of the successor state.
Thus, the prediction of the corresponding expected rewards can be improved leading to a fast convergence towards an optimal policy.

\section{SIMULATION}\label{sec:sim}
In order to examine the RL controller, we use the simplifying assumptions \mbox{$l_{c_1}=l_1$}, \mbox{$l_{c_2}=l_2$}, and the moments of inertia $J_1=m_1 l_1^2$  and $J_2=m_2 l_2^2$.
The basic parameter values of the Acrobot are set to $d_1=0.08$ Nms, $l_1=4$ m, $l_2=1.3$ m, and $m_1=2$ kg for the simulation of the RL controller introduced in Sec.~\ref{sec:reinforcementLearning}. 

We define an initial configuration (ICO) as performance baseline for the following simulation analysis:

\textbf{Performance Baseline Setup ICO} 
\begin{itemize}
    \item The angle $ \theta \in [0^{\circ}, 360^{\circ}] $ is discretized into 36 angular positions with the angular increment being $ \Delta \theta = 10^{\circ} $. 
    The angular velocity $ \dot {\theta} \in [-5.5]\; \unit {rad/s} $ is discretized into 40 elements with step size $ \Delta \dot {\theta} = 0.25 $ rad/s.
    The state space consists of the product of these two sets and an additional terminal state $ s_T $, representing all states outside this set.
    \item The number of episodes in the learning phase is set to 300. Hereby, the episode length is $ \unit[1000]{s} $ using the discretization $ \Delta t_\mathrm{sim} = \unit[10]{ms} $.
    \item The mass of the actuated pendulum is set to  $m_2=2$ kg.
    \item In each time step one of two possible actions $ a \in \mathcal{A} = \{a_1 = $ 'negative angle step of angle $u$', $ a_2 = $ 'positive angle step of angle $u$'~\} can be selected.
    \item We use the reward function Eq.\,\eqref{eq:rewardfunctionstandard}. Hereby, we set the desired energy level of the first pendulum to $ H_d = 1.3 g/l_1 $, which is close to the energy level of the separatrix. Without the second pendulum, the separatrix is located at $H_s=2\alpha$.  
\end{itemize}
\vspace{1mm}




It is of interest for this and other applications with a complicated nonlinear model, whether it is possible for the RL controller to learn and control the corresponding nonlinear dynamical behavior of the Acrobot, given by the equations of motion~\eqref{eq:AcrobotEoM}.

\begin{figure}
    \footnotesize
    \centering	
    \includegraphics[width=1.07 \columnwidth]{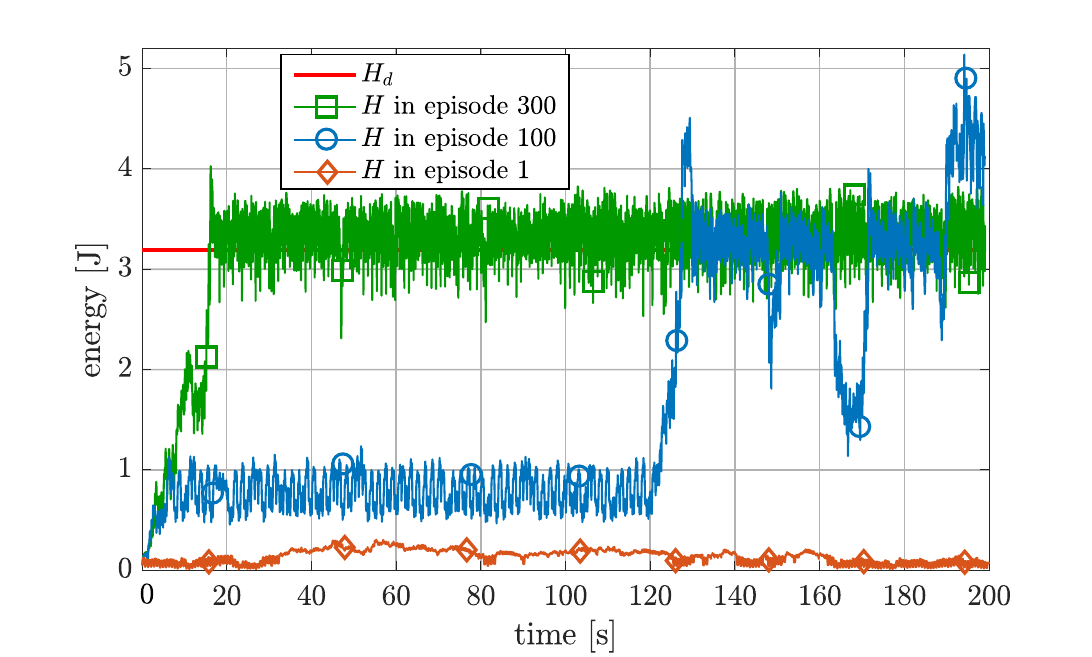}%
    \setlength{\belowcaptionskip}{-10pt}
    \caption{Energy of the first pendulum in episode 1, 100, and 300.}%
    \label{fig:episode1to300}%
\end{figure}
\begin{figure}
    \footnotesize
    \centering	
    \includegraphics[width=1.05\columnwidth]{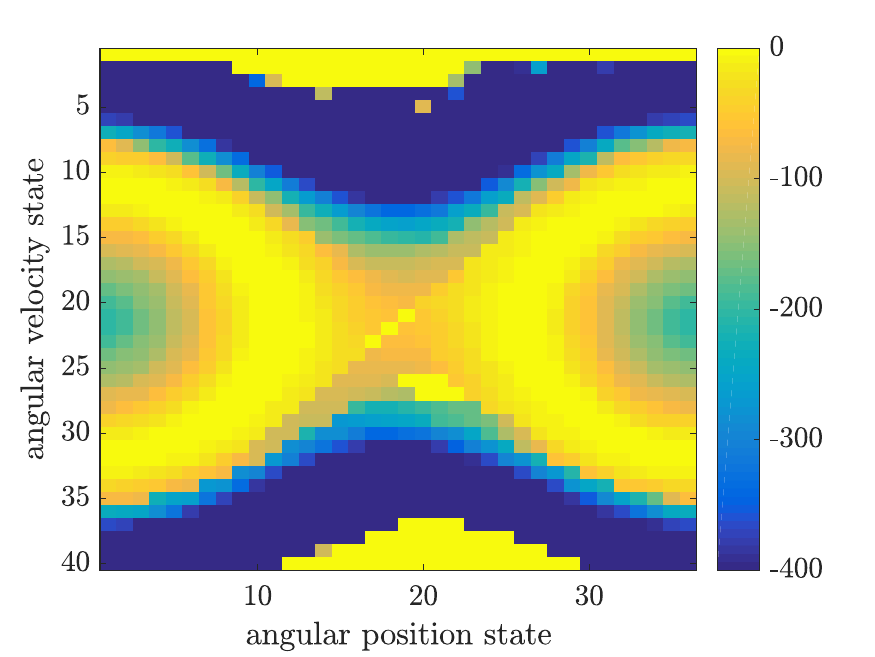}%
    \setlength{\belowcaptionskip}{-10pt}
    \caption{Value function obtained from 300 policy iterations.}%
    \label{fig:valueFunction}%
\end{figure}
\begin{figure}
    \footnotesize
    \centering	
    \input{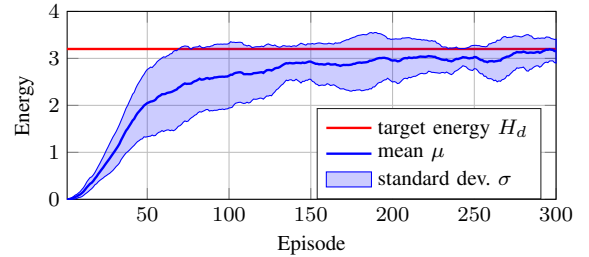}
    \setlength{\belowcaptionskip}{-10pt}
    \caption{Energy $H$ averaged over 10 runs.}%
    \label{fig:EnergyEpisodeSim}%
\end{figure}
\begin{figure}
    \footnotesize
    \centering 						 
    \input{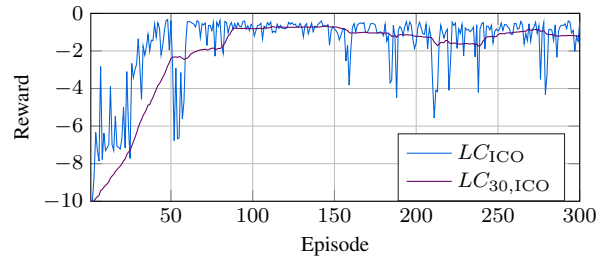}
    \setlength{\belowcaptionskip}{-10pt}
    \caption{Best result for 10 simulation runs using the ICO.}	
    \label{fig:stdconf_best}	
\end{figure}

\textcolor{black}{Simulation results for the ICO show, how the energy evolves in different episodes. This is shown in Fig.~\ref{fig:episode1to300} for episodes 1, 100, and 300, starting at a low randomly chosen initial energy. These results show, that after about 300 episodes the RL~controller learns a policy which maintains the energy of the first pendulum close to a desired energy level. 
The corresponding value function obtained from 300 policy iterations is depicted in Fig.~\ref{fig:valueFunction}. From this, the nonlinear structure of the most valuable states close to the desired energy level $H_d$ can be observed, which corresponds to the expected dynamical behaviour of a pendulum, which is for example described theoretically in}~\cite{dostal2018pendulum}.
\textcolor{black}{In Fig.~\ref{fig:EnergyEpisodeSim} the convergence of the mean energy of the first pendulum towards the desired energy $H_d$ is shown using results from 10 learning runs. This shows, that the RL algorithm converges towards an optimal policy.}
The corresponding ICO learning curve ($LC_{\mathrm{ICO}} $) is depicted in Fig.~\ref{fig:stdconf_best}.
\textcolor{black}{If the trajectories of the first pendulum of the Acrobot leave the angular velocity region $\dot{\theta}\in [-5,5]$ rad/s, than the pendulum is restarted at zero energy. This leads to the non-monotonic convergence of the learning curve shown in Fig.~\ref{fig:stdconf_best}.}
%
%
%
%
%

The learning curve reflects the average reward received according to Eq.\,\eqref{eq:rewardfunctionstandard} within each episode. 
It is obvious, that the learning curve can never reach a value higher than zero. 
The performance of the RL-controller increases with higher values of the learning curve.
Therefore, we will mostly use the learning curves in the following in order to analyze the performance of the various RL-controller setups.

In this example, the values from the $100^\mathrm{th}$ episode onward have small fluctuations with occasional stronger outliers. 
These are often caused by the exploration probability of $ 10\% $, in order that the controller can explore the state space. 
Such outliers and fluctuations of high frequency are obstructive for comparison with other learning curves. 
Therefore, the data is low-pass filtered by forming the so-called moving average over thirty episodes ($ LC_{30} $). 
The result can be seen in Fig.~\ref{fig:stdconf_best}.

\begin{figure}
    \footnotesize
    \centering 						 
    \input{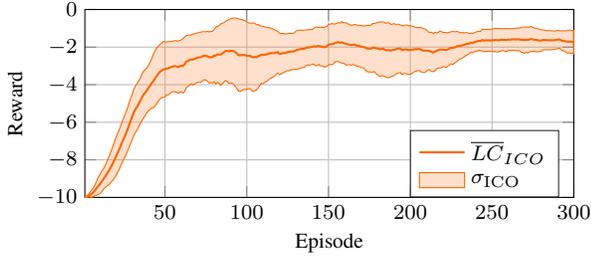}
    \setlength{\belowcaptionskip}{-10pt}
    \caption{The summarized result for 10 independent simulations of the RL controller using the ICO.}
    \label{fig:stdconf}	
\end{figure}

Note that, all learning curves will differ from each other due to their statistical character. 
Therefore, ten simulations are performed and the corresponding mean value ($ \overline{LC}  $), as well as the standard deviation $ \sigma $ are determined. 
The results for the ICO are summarized in Fig.~\ref{fig:stdconf}. \textcolor{black}{There, the corresponding mean value $ \overline{LC}_{\mathrm{ICO}} $  and the standard deviation $ \sigma_{\mathrm{ICO}} $ are shown.} 

Ideally, after a sufficiently long learning period the mean values move towards the zero line, while the standard deviation remains as small as possible. 
In addition, the unfiltered learning curves should not show strong fluctuations.

\section{STUDY AND IMPROVEMENT OF RL CONTROLLER USING SIMULATION}\label{sec:simstudy}
In order to improve the energy stabilizing RL controler for the Acrobot,
we analyze the influence of state space dicretization, modification of the action space, the episode length, and the mass of the actuated pendulum on the performance of the RL controller in the simulation.

\subsection{State space discretization}\label{chap:disk}

One might intuitively assume that learning will be improved when choosing smaller state discretizations. 
However, fine discretizations require larger sets of training data in order to infer a robust policy. 
Moreover, depending on the actual problem the limited memory and processor power might render this option infeasible. 


Let the state space be defined by $ \mathcal{S} $ = $ \mathcal{S}_1 \times \mathcal{S}_2 \cup s_T $, where $ \mathcal{S}_1 $ and $ \mathcal{S}_2 $ contain the discrete angular positions and discrete angular velocities respectively and $s_T$ is the terminal state. 
Both sets $ \mathcal{S}_1 $ and $ \mathcal{S}_2 $ can be either reduced, increased, or left unchanged within the same state limits. 
This corresponds to $ 3^2 $ combinatorial possibilities of changing the state space by an alternative choice of step sizes.
Since there exist also infinite possibilities to choose the sampling intervals, we limit our study to 12 cases which are summarized in Tab.\,\ref{tab:cases1}.
%
%
%
\begin{table}
	\caption{Selected alternative state discretizations.}
	\label{tab:cases1}
	\begin{center}
		\begin{tabular}{|c|c|c|}
			\hline 
			\textbf{Case} & \textbf{State Discretization} & \textbf{States} $ |\mathcal{S}| $\\ 
			\hline 
			ICO & $ \Delta\theta = 10^{\circ}, \Delta\dot{\theta} = \unit[0{,}25]{rad/s} $ & 1441\\			
			\hline 
			1 & $ \Delta\theta = 8^{\circ}, \Delta\dot{\theta} = \unit[0{,}25]{rad/s} $ & 1801\\ 
			\hline 
			2 & $ \Delta\theta = 5^{\circ}, \Delta\dot{\theta} = \unit[0{,}25]{rad/s} $ & 2881\\ 
			\hline 
			3 & $ \Delta\theta = 10^{\circ}, \Delta\dot{\theta} = \unit[0{,}2]{rad/s} $ & 1801\\ 
			\hline 
			4 & $ \Delta\theta = 10^{\circ}, \Delta\dot{\theta} = \unit[0{,}1]{rad/s} $ & 3601\\ 
			\hline 
			5 & $ \Delta\theta = 8^{\circ}, \Delta\dot{\theta} = \unit[0{,}2]{rad/s} $  & 2251\\ 
			\hline
			6 & $ \Delta\theta = 5^{\circ}, \Delta\dot{\theta} = \unit[0{,}1]{rad/s} $  & 7201\\
			\hline
			7 & $ \Delta\theta = 12^{\circ}, \Delta\dot{\theta} = \unit[0{,}3125]{rad/s} $  & 961\\ 
			\hline
			8 & $ \Delta\theta = 20^{\circ}, \Delta\dot{\theta} = \unit[0{,}5]{rad/s} $  & 361\\
			\hline
			9 & $ \Delta\theta = 12^{\circ}, \Delta\dot{\theta} = \unit[0{,}25]{rad/s} $  & 1201\\ 
			\hline
			10 & $ \Delta\theta = 20^{\circ}, \Delta\dot{\theta} = \unit[0{,}25]{rad/s} $  & 721\\
			\hline
			11 & $ \Delta\theta = 10^{\circ}, \Delta\dot{\theta} = \unit[0{,}3125]{rad/s} $  & 1153\\ 
			\hline
			12 & $ \Delta\theta = 10^{\circ}, \Delta\dot{\theta} = \unit[0{,}5]{rad/s} $  & 721\\
			\hline
		\end{tabular}
	\end{center}
\end{table}
%
%
The results for these cases are depicted in Fig.~\ref{fig:posfein}. 
Note that the last index denotes the corresponding case number, such as $LC_{30,1}$ for the learning curve of case 1, which is averaged over 30 episodes.
Reducing the step size from $ 10^{\circ}$ to $ 8^{\circ}$ qualitatively worsens the result, \textcolor{black}{as depicted in Fig.\,\ref{fig:case123_posfein} for case 1,} with respect to the ICO-baseline from Fig.~\ref{fig:stdconf}. 
Smaller rewards are achieved and the average values fluctuate comparatively strongly. 
Halving the step size to $\Delta\theta = 5^{\circ}$ reduces the reward even more, \textcolor{black}{as can be seen from the curve of case 1 in Fig.\,\ref{fig:case123_posfein}.} 



\textcolor{black}{When considering Fig.\,\ref{fig:case456_velfein},} the red curve for case 4 reaches a reward value of $ -4 $ and the large standard deviation is clearly visible.
The curve for case 3, on the other hand, performs much better between episodes 100 and 200. 
The standard deviation gradually decreases and the curve of the corresponding mean moves into the interval $ [-2, -1] $. 
For cases 5 and 6 \textcolor{black}{in Fig.\,\ref{fig:case456_velfein},} we reduce the discretization for both states,
which leads to a less pronounced initial learning phase. In addition, the mean values become smaller and smaller with increasing refinement of the state space.
Apparently the state space cannot be sufficiently explored due to its size. As a consequence, the quality of the learned policy decreases. 

\begin{figure}
\footnotesize
	\centering 						 
	\begin{subfigure}[b]{0.9\columnwidth}
     \centering
     \input{posfein.tikz}
     \caption{Fine position state discretization.}
     \label{fig:case123_posfein}
    \end{subfigure}
    \begin{subfigure}[b]{0.9\columnwidth}
         \centering
         \input{velfein.tikz}
         \caption{Fine velocity state discretization.}
         \label{fig:case456_velfein}
    \end{subfigure}
    \begin{subfigure}[b]{0.9\columnwidth}
         \centering
         	\input{posvelgrob.tikz}
         	\caption{Coarse position and velocity state discretization.}
         	\label{fig:case789_posvelgrob}
    \end{subfigure}
    \begin{subfigure}[b]{0.9\columnwidth}
         \centering
         \input{posgrob.tikz}
         \caption{Coarse position state discretization.}
         \label{fig:case101112_posgrob}
    \end{subfigure}
	\caption{Summarized results for cases 1 to 12, according to the case definitions in Tab.\,\ref{tab:cases1}.}	
	\label{fig:posfein}	
\end{figure}

In cases 7 and 8  we consider the scenario of a slight increase of $ \Delta\theta $ and $ \Delta\dot{\theta} $, \textcolor{black}{as displayed in Fig.\,\ref{fig:case789_posvelgrob}.} 
After about 100 episodes, the curve for case 7 lies mostly below the value of $ -2 $ and gets lower than the curve for case 8 towards the end. 
Furthermore, the graph for case 7 shows a notable behaviour of the standard deviation, as it increases strongly towards the end.
Case 8 considers a strong reduction of the state space from 1441 to 361 states. 
For this case, large fluctuations in the first half of the learning period can be observed. Nevertheless, these fluctuations are reduced and the mean value approaches the zero line. 

\begin{figure}
	\centering 						 
	\input{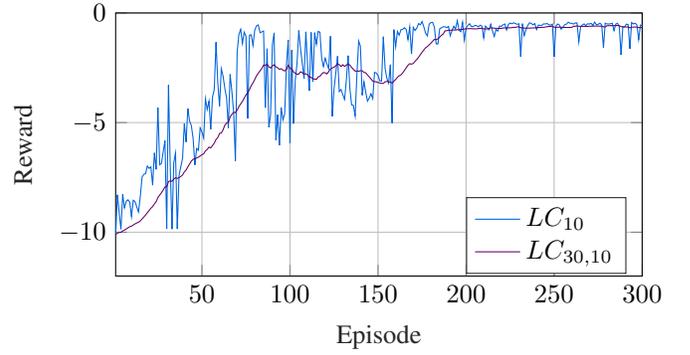}
	\caption{Best learning curve for case 11.}			
	\label{fig:posgrobgrobbest}	
\end{figure}

From cases 9 and 10, \textcolor{black}{which are shown in Figs.\,\ref{fig:case789_posvelgrob} and \ref{fig:case101112_posgrob},} it can be seen that changes of the
step size $ \Delta\theta $ affect the results only slightly, since the corresponding mean values, as well as the  
standard deviations are quite similar. 
The mean values remain within the interval $ [-4,-2] $ and the standard deviations increase towards the end. 
%
Case 11 and 12 led to the 
best obtained learning curves in the simulations, \textcolor{black}{as shown in Fig.\,\ref{fig:case101112_posgrob}.} One of these learning curves is shown in Fig.~\ref{fig:posgrobgrobbest}. There, the values remain near zero from episode 200 with smallest fluctuations. Apart from a few small peaks, the original curve almost coincides with the sliding average. 
In comparison to the baseline ICO configuration, the scattering at the beginning of the learning phase is higher. 
The curves for cases 11 and 12 have an almost identical quality. 
The doubling of the step size even caused an improvement.
Therefore, this case was also tested in the experiment, as can be seen in section~\ref{sec:EXPERIMENTAL_RESULTS}. 



In summary, cases 9-12 show that, although the RL algorithm faced a strong reduction of the state space, a good control policies can still be found. 
In contrast, the analysis of setups with finer discretization showed rather negative tendencies except for case 3. 
However, the result might differ when longer episodes are considered. 

\subsection{Length of the episodes}\label{chap:el}

The longer the episode length the more data is available to infer a control policy from.
On an intuitive basis, the controller requires enough time to, first, explore the system dynamics and than, second, exploit this knowledge within its control policy. 
In order to identify trends, we run simulations using the test cases from Tab.~\ref{tab:cases_episode}. 


Note that the total learning time remains constant with a total of $ \unit[300000]{s} $. 
As a result, the total number of episodes and thus the number of strategy optimizations varies. 

\begin{figure}[h]
	\centering 						 
	\input{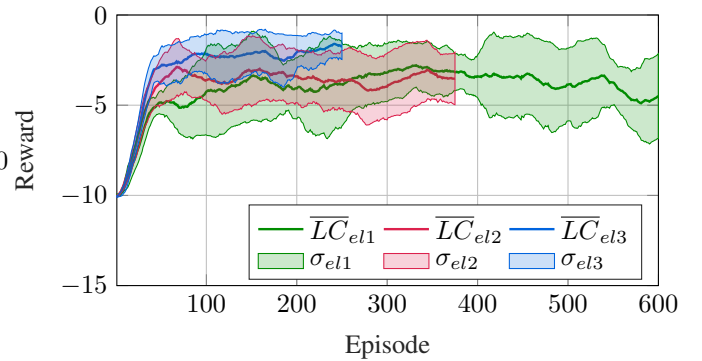}
	\caption{Summarized results: Case \textit{el1} to case \textit{el3}.}			
	\label{fig:episodes}	
\end{figure}

\begin{figure}
	\centering 						 
\input{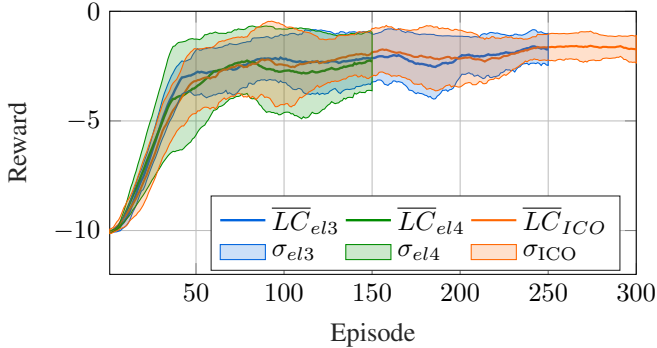}
	\caption{Summarized results: Case \textit{el3}, case \textit{el4}, and \textit{ICO}.}			
	\label{fig:episodes2}	
\end{figure}

Fig.~\ref{fig:episodes} illustrates the development when the episode length is changed according to Tab.~\ref{tab:cases_episode}. 
 \begin{table}
	\caption{Selected alternatives for changing the episode length.}
	\label{tab:cases_episode}
	\begin{center}
		\begin{tabular}{|c|c|}
			\hline 
			\textbf{case} & \textbf{time steps} \\
			\hline 
			\textit{ICO} & 100,000 \\ 			
			\hline 
			\textit{el1} & 50,000 \\ 
			\hline 
			\textit{el2} & 80,000  \\ 
			\hline 
			\textit{el3} & 120,000  \\ 
			\hline 
			\textit{el4} & 200,000 \\ 
			\hline 
		\end{tabular}
	\end{center}
\end{table}
%
%
%
\textcolor{black}{It can be seen that} for the case $el1$ with halved number of time steps compared to the ICO, the mean values are lower and show a more irregular behavior. This is due to the fact that the training of the swing-up phase dominates in this case, but keeping the energy at the desired energy level is less trained. 
If the episode length is shortened by $ 20,000 $ time steps (case \textit{el2}), the mean value still remains below $-2$, but the curve shows less variation. 
The overall result improves not significantly if the episode length of the ICO is extended by $ 20,000 $ time steps, which is case $el3$, as can be observed in Fig.~\ref{fig:episodes2}. 
%
%
Thus, we see that after the learning period's half-time the average reward shows a relatively smooth behaviour. Moreover, there are no more significant outliers. 
From episode 80 onwards, the moving average runs almost horizontally and close to the zero target line. 
This means that including additional episodes will probably only smooth the original curve. 

However, the overall results including the cases \textit{el3} and \textit{el4} are worse than the ICO-baseline, as can be seen in Fig.~\ref{fig:episodes2}. Since the total learning time remains the same, the controller can only swing up $150 $ times in case \textit{el4} or $250 $ times in case \textit{el3} instead of $300 $ times compared to the ICO and the swing-up strategy is therefore less often trained. 

In conclusion longer episodes result in a relatively longer time spent within the stabilization phase. In contrast, short episodes support learning of the swing-up phase. In the ICO setup these phases are already well balanced.
 
\subsection{Augmented action space}\label{chap:action}

So far, the RL algorithm only considers two actions.
While an actuation with a larger angular amplitude of the second pendulum is advantageous in the swing-up phase, a small amplitude or even no action might be beneficial to stabilize the Acrobot at the desired energy level. 
Therefore,  the action space is augmented by additional action states in order to explore possible positive impacts on the optimized control policy.

We introduce the following actions:  $ a_3 =$ 'do nothing', $ a_1 =$ 'negative angle step', $ a_2 =$ 'positive angle step', $ a_4 = $ 'minimum angle', $ a_5 =$ 'maximum angle'.
An early testing showed that an augmentation with all five actions at the same time considerably decreases the control performance. 
This is likely because the RL controller is not able to explore the Acrobot dynamics sufficiently, as it faces too many possible actions for the given constant learning time. 
Due to the increase of action states, the transition matrix $ \mathbf{P}(s'|s,a) $ increases and thus the training data needed to learn the state transition sufficiently. 
As a result, the size of the action space should remain as small as possible.
Hence, we examine the following reduced action space configurations: $ \mathcal{A}_1= $\{$ a_4,a_5 $\}, $ \mathcal{A}_2= $\{$ a_3,a_4,a_5 $\} and $ \mathcal{A}_3= $\{$ a_1,a_2, a_3$\}. 

Note that the action space remains the same in both phases, the swing-up and the stabilization phase. 
This enforces that the finally selected action set serves as a good compromise for both control phases. 

\begin{figure}
	\centering 						 
	\input{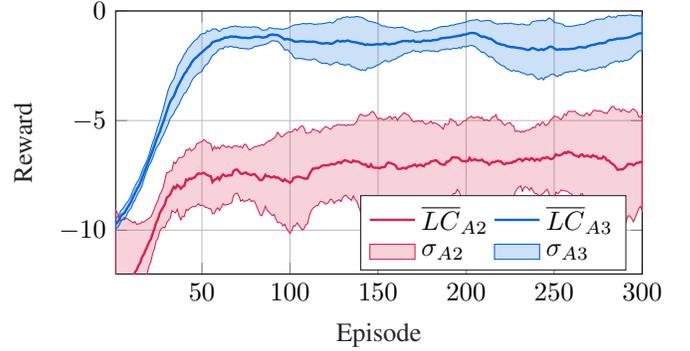}
	\caption{Learning curves for the action sets $ \mathcal{A}_2= $\{$ a_1,a_2,a_3 $\} and $ \mathcal{A}_3= $\{$ a_3,a_4,a_5 $\}.}			
	\label{fig:A2}	
\end{figure}

The simulation runs using action set $ \mathcal{A}_1 $ do not achieve satisfactory performance.
The controller can not keep the deviation from the target value within satisfactory limits. Using the set $ \mathcal{A}_2 $ and $ \mathcal{A}3 $ including the action $ a_3 =$ 'do nothing', the learning curves depicted in Fig.~\ref{fig:A2} were achieved. 
%
\textcolor{black}{As can be seen in this figure, the learning process is considerably worse compared to the use of the ICO, if set $ \mathcal{A}_2 $ is used.} 
For the case $ \mathcal{A}_3 $, which is when the ICO action space is expanded with action $ a_3 $, a positive effect on the learning curve is observed.
%
%
The $\overline{LC}_{A3}$-curve is above the value -2 while its standard deviation is comparatively small.
Compared to the learning curve from Fig.~\ref{fig:stdconf_best} the initial learning phase can be seen as already completed after about 40 episodes.

\subsection{Mass of the actuated pendulum.}\label{chap:mm}



The servo actuator of the second pendulum influences the kinetic and potential energy of the overall system. 
In the experiment, the servo motor has a limited actuating torque, which can be applied in an on/off-fashion. 
Thus, it is the maximum for each action. 
If the mass is increased, the maximum angular acceleration $ \ddot{u} $ decreases accordingly due to the mass inertia. 
In the simulation, however, the dynamics of the motor are simplified resulting in modelling error. 
The angular velocity $ \dot{u} $ is assumed to be constant and independent of $ m_2 $. 
If the  mass $ m_2 $ is reduced, for example, each action supplies less energy per time step to the system.
Note that, especially in the initial swing-up phase, high energy supply is required to reach the desired energy level $H_d$ as quickly as possible. 
However, a high mass in combination with the actions $ a_1 =$ 'negative angle step' and $ a_2 =$ 'positive angle step' can be challenging when stabilizing the desired energy level, which requires sensitive energy input. Thus, a trade-off between objectives arises. 

\begin{figure}
	\centering 						 
\input{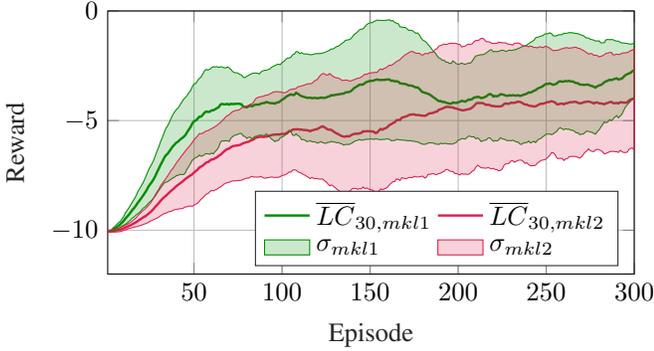}
	\caption{Control with smaller mass.}			
	\label{fig:masseklein}	
\end{figure}

In a first step, we reduce the mass $ m_2 $ to $ \unit[1{,}5]{kg} $ $ (LC_{mkl1}) $ and in a second step to $ \unit[1]{kg} $ $ (LC_{mkl2}) $ for the simulation. 
The simulation results are summarized in Fig.~\ref{fig:masseklein}. 

%
\textcolor{black}{It can be seen in this figure that} in comparison to the ICP-baseline, the achieved rewards are comparatively small and widely spread.
In addition, the slope of the curves in the initial learning phase decreases with the reduction in mass. 
Resulting in an overall worse result caused by the poorer swinging capability of the second pendulum. 
Hence, the swing-up phase therefore requires more time. 
As a result, less reward is achieved which affects the mean value accordingly. 

\begin{figure}
	\centering 						 
\input{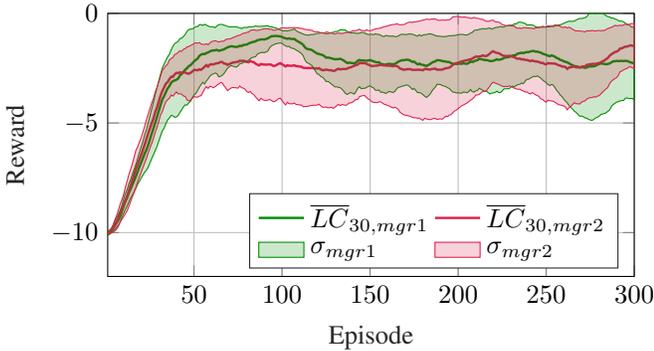}
	\caption{Control with larger mass.}			
	\label{fig:massegross}	
\end{figure}

\begin{figure}
	\centering 						 
    \input{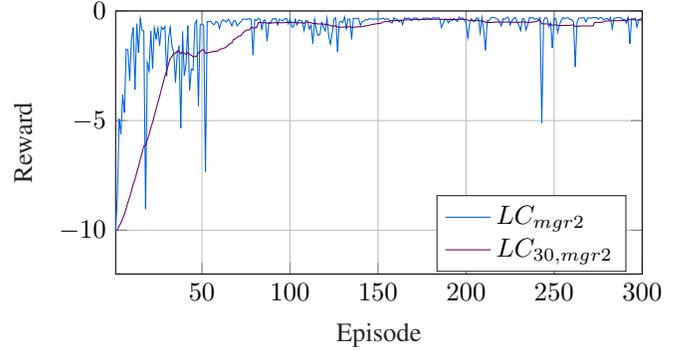}
	\caption{Best learning curve obtained after doubling the mass.}			
	\label{fig:massegross_best}	
\end{figure}

This is very different to the setup where the mass is, first increased to $ \unit[3]{kg} $ $ (LC_{mgr1}) $ and then even further to $ \unit[4]{kg}$ $ (LC_{mgr2}) $, as shown in Fig.~\ref{fig:massegross}. 
%
\textcolor{black}{There it can be observed that the mean value curves are again in the same order of magnitude as the ICO-baseline.} 
This is because the first pendulum gains more momentum from the movement of the second pendulum. 
At the same time, the ability to stabilize the pendulum is slightly reduced. 
The initial learning phase (\textless episode 50) is also clearly visible.
In addition, the difference between the case $LC_{mgr1}$ and $LC_{mgr2}$ is relatively small. 
The mass doubling also results in a very good learning curve, which is shown in Fig.~\ref{fig:massegross_best}.
%
%
\textcolor{black}{There,} we can observe that already after 50 episodes very high rewards are received. 
Apart from some peaks towards the end of the learning period, the learning curve is relatively smooth and the sliding average is very close to the zero target value.


Finally, we can conclude from the test scenarios above that the functionality of the RL algorithm seems to be independent from the mass.
Moreover, the physical parameters, such as the mass or size of the angular steps, determine how high the rewards will be.
This is because the RL algorithm de-facto only uses the means in order to achieve the desired target.  
Consequently, the mass, like other physical parameters, has to be chosen in such a way that it is compatible with the desired dynamical behaviour before the actual learning process is started. 

\section{EXPERIMENTAL SETUP}
\textcolor{black}{In this section we demonstrate the performance and robustness of our method using the proposed RL algorithm in an Acrobot experimental setup, which has only limited computational resources. The experimental setup can be seen in Fig.~\ref{fig:doublependulum}b, where also a link to an accompanying video can be found.}
%
%
\textcolor{black}{The computations are conducted onboard a RaspberryPi\,3B+ single board computer (CPU: $4\times1.4$\,GHz, RAM: 1\,GB), which is light enough to be mounted on the Acrobot's flywheel. }
Note that our RL algorithm is currently implemented in Python. 
We use a low-cost servo motor to drive the motion of the second pendulum. 
The Acrobot's angular position and velocity are determined based on the IMU data for which we choose a MPU-6050 GyS21 as a sensor.
Because the gyroscope is prone to drift, we compensate the bias of current gyroscope reading based on the initial gyroscope reading. 
Moreover, we use a complementary filter in order to accurately determine the current angular position of the pendulum. 
The filtered measurements of the angular position $\theta$ and velocity $\dot{\theta}$ are then sent to the algorithm. 
The Acrobot is a rotary system. 
Hence, the power supply has to be realized on the flywheel which is done through standard USB-powerbanks. 
The overall mounting and wiring of the components can also be seen in Fig.~\ref{fig:doublependulum}b.
Note that all additional mountings have an impact on the Acrobot dynamics and posing additional challenges on the RL-algorithm.


We want to examine the performance for learning from scratch. Thus, no knowledge of the equation of motion is needed to apply the RL controller in the experiments. This makes the use of RL control especially appealing for deployment in scenarios which are complex to model.

We use the following initial configuration for the experimental setup ($\mathrm{ICO}_{\mathrm{exp}}$) in order to examine the performance and robustness of various setups:  

\textbf{Performance Baseline Setup} $\mathbf{ICO}_{\mathrm{exp}} $
\begin{itemize}
    \item We choose the same state space discretization as for the simulation in Sec.~\ref{sec:sim}.
    \item The real experiments require much more time than the simulation. 
    Therefore, the length of an episode is reduced to $ \unit[40]{s} $ and the total learning time to $ 100 $ episodes.
    In contrast to simulation the real servo actuator requires time for positioning. Thus, we time step to $ \Delta t_{\mathrm{exp}} =\unit[0{.}1]{s} $.
    \item Mass of the actuated pendulum is set up as stated in Tab.~{tab:params}: tip mass $ m_p = \unit[0{.}038]{kg} $, rod mass $ m_r = \unit[0{.}042]{kg}$.  
    \item Analogue to the simulation the algorithm can choose between two actions $ a \in \mathcal{A} = \{a_1 = $ 'negative angle step', $ a_2 = $ 'positive angle step'~\}. 
    Thereby,  we define the angle step as the maximal possible angle step which can be achieved by the servo motor in the time step $ \Delta t_{\mathrm{exp}} =\unit[0{.}1]{s} $, applying the maximal available torque of the servo motor.
    Moreover, the angle $ u $ is restricted to the interval $ [90^{\circ}, 270^{\circ}] $.
    \item  For the experiments, the reward function from Eq.\,\eqref{eq:scaledhamiltonian} is used. 
    The desired energy level $ H_d $ is scaled in the experiment and is initialized by $ \widetilde{H_d} = \unit[4{,}4]{Nm} $.
\end{itemize}
\vspace{1mm}

In order to ensure that the controller also learns well in the swing-up phase, a braking function is implemented since the system's damping is relatively small. For this,the second pendulum is used as an active damper in order to drive the first pendulum to its rest position.
This allows the controller to begin the next episode from a stable rest position where $ \widetilde{H} \approx 0 $.
This drastically reduces the manual effort for collecting data. 

\begin{figure}
    \centering 			
    \footnotesize			 
    \input{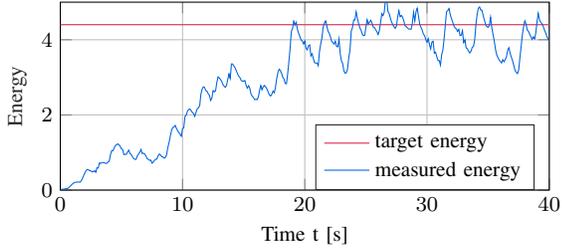}
    \setlength{\belowcaptionskip}{-10pt}
    \caption{Energy evolution in the 100th episode using the ICO.}			
    \label{fig:exp_energy}
\end{figure}

In Fig.~\ref{fig:exp_energy} an example of the energy curve after 100 episodes is shown for the proposed $\mathrm{ICO_{exp}}$. \textcolor{black}{The individual episodes last 40-80\,s before the termination criteria are met.}

\section{EXPERIMENTAL RESULTS}\label{sec:EXPERIMENTAL_RESULTS}
In this section we present a set of experimental results for the RL controller of the Acrobot, whereby the case of energy stabilization at a desired level and for the control of rotation is considered.

\subsection{Energy Level Control}\label{sec:energycontrol}
In the first part of this section we examine the RL controller's performance and robustness on energy level stabilization. 
For this, we analyze the influence of state space discretization, modification of the action space, the episode length, and the mass of the actuated pendulum on the performance of the RL controller in the experiment. 
As described in Section~\ref{sec:simstudy}, we have run several simulations using modifications of the ICO from section~\ref{sec:sim} beforehand, in order to figure out meaningful setups for our experimental study. This saves a significant amount of time, since learning in the simulation is much faster than in the experiment. It turned out for example, that good learning performance can also be achieved with the coarser discretization of the state space. Better results were achieved with twice the step size $ \Delta\theta $. From these studies, we have chosen the best candidates for the set-up modifications in the experiment, which are summarized in the following.




In order to account for the performance of the RL controller in the swing-up phase and the hold phase, where the desired energy level has to be maintained, the corresponding learning curves are considered separately.
Starting from results of the $\mathrm{ICO_{exp}}$ baseline as shown in Fig.~\ref{fig:exp_energy}, each episode is split into the time interval $ [0,20]\unit{s} $, which is the \textit{swing-up} phase, and $ [20,40]\unit{s} $, which is the \textit{hold} phase. \textcolor{black}{Thereby the \textit{swing-up} phase consists of the part, where the energy curve starts at a low value, whereas the \textit{hold} phase consists of the part of the energy curve which lies in a regionclose to the target energy level, cf. Fig.~\ref{fig:exp_energy}.} This split results in two learning curves for each experimental run instead of one. 
In order to combine the information from two experiments, the mean value $ \overline{LC}_\mathrm{exp} $ of two independent learning-curves $ LC_\mathrm{exp} $ is determined. 
This split allows for a more detailed analysis of the parameter influence.

\begin{figure}
    \footnotesize
    \centering 						 
    \input{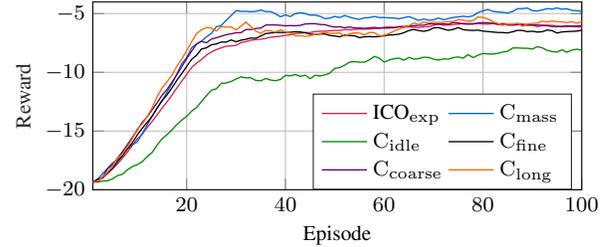}
    \setlength{\belowcaptionskip}{-10pt}
    \caption{Mean reward in the swing-up phase.}			
    \label{fig:comparison1}	
\end{figure}

The examined Acrobot configurations are summarized in Tab.\,\ref{tab:cases}.
\begin{table}
    \caption{Examined Acrobot configurations.}
    \label{tab:cases}
    \begin{center}
        \begin{tabular}{|c|c|c|}
            \hline 
            \textbf{Case} & \textbf{Parameterization} & \textbf{\# States} $ |\mathcal{S}| $\\ 
            \hline 
            ICO${}_\mathrm{exp}$ & $ \Delta\theta = 10^{\circ}, \Delta\dot{\theta} = \unit[0{.}25]{rad/s} $ & 1441\\			
            \hline 
            $\mathrm{C}_\mathrm{fine}$ & $ \Delta\theta = 10^{\circ}, \Delta\dot{\theta} = \unit[0{.}20]{rad/s} $ & 1801\\ 
            \hline 
            $\mathrm{C}_\mathrm{coarse}$ & $ \Delta\theta = 20^{\circ}, \Delta\dot{\theta} = \unit[0{.}25]{rad/s} $ & 721\\ 
            \hline 
            $\mathrm{C}_\mathrm{idle}$ &  ICO with $\mathcal{A}_\mathrm{extended} = \{a_1, a_2, a_\mathrm{idle}\}$  & 1441\\ 
            \hline 
            $\mathrm{C}_\mathrm{long}$ &  ICO with doubled episode length  & 1441\\ 
            \hline 
            $\mathrm{C}_\mathrm{mass}$ & ICO with increased pendulum mass $m_p$   & 1441\\ 
            \hline
        \end{tabular}
    \end{center}
\end{table}
Using these configurations, the corresponding learning curves for the swing-up phase are shown in
Fig.~\ref{fig:comparison1}. 
%
\textcolor{black}{It can be seen in this figure that} with a larger pendulum mass $\mathrm{C}_\mathrm{mass}$ the highest reward is generated. 
Consequently, the performance of the \textit{swing-up control} of the first pendulum is best for the considered cases. 
Applying the coarse discretization $\mathrm{C}_\mathrm{coarse}$, the learning curve is steeper than that of the ICO${}_\mathrm{exp}$. 
Subsequently, they overlap and strive towards the same value. 
A doubled  episode length $\mathrm{C}_\mathrm{long}$ leads to a lower reward for the same learning duration, as can be seen towards the end of the 50th episode. 
The same applies to the fine discretization $\mathrm{C}_\mathrm{fine}$. 
In addition, the worst performance during the swing-up was achieved when using the extended action space in case $\mathrm{C}_\mathrm{idle}$.

\begin{figure}
    \footnotesize
    \centering 						 
    \input{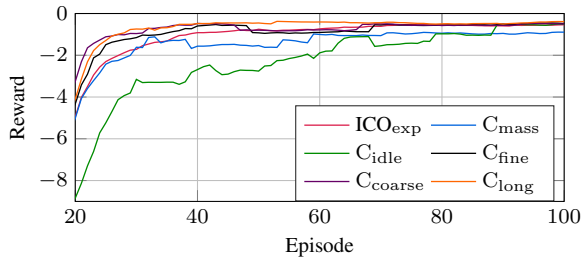}
    \setlength{\belowcaptionskip}{-10pt}
    \caption{Mean reward in the hold phase}
    \label{fig:hold}	
\end{figure}

Analogue, the learning development during the \textit{hold phase} is shown in Fig.~\ref{fig:hold}. 
\textcolor{black}{In this figure it can be observed that} the stabilization performance decreases with increased pendulum mass $\mathrm{C}_\mathrm{mass}$. 
Moreover, the use of longer episodes $\mathrm{C}_\mathrm{long}$ leads to good results after only 50 episodes, which are at least as good as with the ICO${}_\mathrm{exp}$. 
Using the discretization cases $\mathrm{C}_\mathrm{fine}$ and $\mathrm{C}_\mathrm{coarse}$, learning of the RL controller shows a better performance, as the learning curves are higher than the learning curve of the ICO${}_\mathrm{exp}$. However, towards the end of learning time they reach the same value, which means that the learning algorithm adapted well to the different situations.
As far as the action space $ \mathcal{A}_3 $ is concerned, relatively many episodes pass until finally the level of the ICO${}_\mathrm{exp}$is reached. 

Although using a greater mass results in higher reward during the swing-up phase, the hold phase reward is insensible to the chosen set-up variations as they converge to almost the same reward values towards the end of the learning time. 
This can be clearly seen in Fig.~\ref{fig:hold}, where many differences in the beginning of the learning period can be observed.
The learning time is therefore well utilized by the RL algorithm, regardless of the studied different circumstances and setups. 
For the case of the doubled angular step size $\Delta\theta=20^\circ$ in $\mathrm{C}_\mathrm{coarse}$ learning is accelerated without noticeably affecting the control performance. 
As a result, this configuration seems promising direction for resource savings in future work.

Based on the previous results, we now examine combinations of the above setups to explore whether performance can be improved even further
Therefore, we combine the double angular increment  $\mathrm{C}_\mathrm{coarse}$  with longer episodes  $\mathrm{C}_\mathrm{long}$, which both achieved promising results according Fig.~\ref{fig:hold}. 
Moreover, we examine in a series of ten experimental cycles for the setup  $\mathrm{C}_\mathrm{coarse,long}$ whether extending the action space by the \textit{idle}-action $a_\mathrm{idle}$ has an influence on the performance.
This experiment intended to show whether slow learning due to the idle action can be compensated with the advantages of double step size and longer episodes.
%
In Fig.~\ref{fig:kombi1} and Fig.~\ref{fig:kombi2} the results for the swing-up phase and the hold phase are shown, respectively. 
\begin{figure}[h]
    \footnotesize
    \centering 						 
%
%
%
\begin{tikzpicture}

\begin{axis}[%
width=7.1cm,
height=4cm,
scale only axis,
xmin=1,
xmax=100,
xlabel style={font=\color{white!15!black}},
xlabel={Episode},
ymin=-20,
ymax=-5,
ylabel style={font=\color{white!15!black}},
ylabel={Reward},
xlabel near ticks,
ylabel near ticks,
axis background/.style={fill=white},
xmajorgrids,
ymajorgrids,
legend style={at={(0.98,0)},anchor=south east,legend cell align=left,align=left,draw=white!15!black},
]
\addplot [color=mumred, mark=square, mark repeat=20, line width=0.5pt]
  table[row sep=crcr]{%
	1	-19.36\\
	2	-19.1934165612248\\
	3	-18.8877603908529\\
	4	-18.6288029331398\\
	5	-18.212324820004\\
	6	-17.800910114164\\
	7	-17.3028356502931\\
	8	-16.8092215581595\\
	9	-16.2220777913807\\
	10	-15.6660036545702\\
	11	-15.2827603768171\\
	12	-14.7159258489112\\
	13	-14.1981007401114\\
	14	-13.6309572584872\\
	15	-13.0542857039907\\
	16	-12.5324252849627\\
	17	-11.9544367834922\\
	18	-11.3409638990731\\
	19	-10.7492703884568\\
	20	-10.1838235394787\\
	21	-9.58658765431943\\
	22	-9.22549813149089\\
	23	-8.91081648886352\\
	24	-8.53770987973604\\
	25	-8.26744610460834\\
	26	-8.06925394682114\\
	27	-7.91563629371282\\
	28	-7.77758662626454\\
	29	-7.75022196703644\\
	30	-7.71506313517369\\
	31	-7.46346258353405\\
	32	-7.39551593960493\\
	33	-7.26601044379254\\
	34	-7.22239076485364\\
	35	-7.18232775690552\\
	36	-7.11651820025483\\
	37	-7.01229126887782\\
	39	-6.92449120195043\\
	40	-6.85344743175057\\
	41	-6.8043172787083\\
	42	-6.70133161052097\\
	43	-6.66506670778175\\
	44	-6.6201800450725\\
	45	-6.61769805534385\\
	46	-6.53548267525659\\
	47	-6.50937564700371\\
	48	-6.49742346070268\\
	49	-6.46407819341471\\
	50	-6.43673318903687\\
	51	-6.43959230105722\\
	52	-6.36201003345637\\
	53	-6.38601370362996\\
	54	-6.34093739157426\\
	56	-6.31780426387732\\
	57	-6.31705996170062\\
	59	-6.24681758711392\\
	60	-6.16110686140723\\
	61	-6.13632109364481\\
	62	-6.16710913142461\\
	63	-6.13289495943575\\
	64	-6.16537767483729\\
	65	-6.12773697903947\\
	66	-6.16729267957767\\
	67	-6.19371606521018\\
	68	-6.13595048669926\\
	69	-6.11268293517492\\
	70	-6.00521372096019\\
	71	-5.86520726873739\\
	72	-5.99847472973909\\
	73	-5.97118248757832\\
	75	-5.90401573190718\\
	76	-5.83516950928633\\
	77	-5.84063824241602\\
	78	-5.85609661557534\\
	79	-5.85169734259787\\
	80	-5.88958143699125\\
	81	-5.90459070897477\\
	82	-5.84892430275117\\
	83	-5.89999608447479\\
	84	-5.84351802062253\\
	85	-5.89070108123447\\
	86	-5.90296847120206\\
	87	-5.85909684109146\\
	88	-5.89830216283787\\
	89	-5.91903821268453\\
	90	-5.96869361089414\\
	91	-6.1128807706547\\
	92	-6.02064444333705\\
	93	-5.99276746099356\\
	94	-6.00248101888558\\
	95	-6.07118125697751\\
	96	-6.11932227567249\\
	97	-6.12056383865459\\
	98	-6.09339614232606\\
	99	-6.13153815189732\\
	100	-6.13750914104807\\
};

\addplot [color=mumblue, mark=o, mark repeat=20,  line width=0.5pt]
  table[row sep=crcr]{%
	1	-19.36\\
	2	-19.3386534406835\\
	3	-19.2237633360214\\
	4	-19.003249292729\\
	5	-18.8986378140276\\
	6	-18.8699670617872\\
	7	-18.5601834269238\\
	8	-18.2273189161062\\
	9	-17.8560105932102\\
	10	-17.5047318637766\\
	11	-16.9392490258185\\
	13	-15.7514629588426\\
	14	-15.1372519206271\\
	15	-14.6850211769077\\
	16	-14.1828859829311\\
	17	-13.6944952536943\\
	18	-13.1894496091655\\
	21	-11.412603048529\\
	22	-10.773432940731\\
	23	-10.2375452909989\\
	24	-9.8326017085476\\
	25	-9.31344595900116\\
	26	-8.70831375499645\\
	27	-8.33354193250486\\
	29	-7.82334847848841\\
	30	-7.50650747014085\\
	31	-7.45258445430538\\
	32	-7.46885646098309\\
	33	-7.28074129870484\\
	34	-7.25997931331354\\
	35	-7.00600623693322\\
	36	-6.89826688606719\\
	37	-6.76412592284697\\
	38	-6.66843501631648\\
	39	-6.58543831896151\\
	40	-6.57595246494499\\
	41	-6.70618107327405\\
	42	-6.88168077530283\\
	43	-6.8157884955163\\
	44	-6.7951872499804\\
	45	-6.74537416200445\\
	46	-6.78060130313226\\
	47	-6.94408245694825\\
	48	-7.0524251727249\\
	49	-7.06913634145506\\
	50	-7.04999067261024\\
	51	-7.0026421462105\\
	52	-6.96960754742791\\
	53	-7.16695440904668\\
	54	-7.30607106072759\\
	55	-7.42870495999303\\
	56	-7.39108553637045\\
	57	-7.40262321040132\\
	58	-7.3613537394495\\
	59	-7.53242427839221\\
	60	-7.55278526822747\\
	61	-7.39445091951647\\
	62	-7.25670302036313\\
	63	-7.31076116977108\\
	64	-7.15277557139778\\
	65	-7.20308686626653\\
	66	-7.12116757280782\\
	67	-6.94501698681333\\
	68	-6.8085459727909\\
	69	-6.75038233936874\\
	70	-6.83854312282652\\
	71	-6.85780183230321\\
	72	-6.81695583987899\\
	73	-6.8085205400135\\
	74	-6.67932453496807\\
	75	-6.59944979404038\\
	76	-6.60299178000228\\
	77	-6.6478725925455\\
	78	-6.66286100918535\\
	79	-6.53043029347721\\
	80	-6.44274008033813\\
	81	-6.42294003822705\\
	82	-6.46040926779662\\
	83	-6.44743928720774\\
	84	-6.54462425302353\\
	85	-6.61828574003823\\
	86	-6.63694548954328\\
	87	-6.6193016705565\\
	88	-6.6097429403949\\
	89	-6.68546665211278\\
	90	-6.64599701829316\\
	91	-6.67224794934566\\
	92	-6.6673155045971\\
	93	-6.62359866229674\\
	94	-6.59618197885143\\
	95	-6.64575304394519\\
	96	-6.6320711462362\\
	97	-6.50477502123356\\
	98	-6.52736973109364\\
	100	-6.42778191656072\\
};

\addplot [color=mumgreen, mark=diamond, mark repeat=20,  line width=0.5pt]
table[row sep=crcr]{%
	1	-19.36\\
	2	-19.2437113286104\\
	3	-19.0312387693183\\
	4	-18.7363306681343\\
	5	-18.381164660642\\
	6	-18.0404264316435\\
	7	-17.5756860585731\\
	8	-17.0124821624194\\
	9	-16.4211831076456\\
	11	-15.2512993005037\\
	12	-14.5678382824267\\
	13	-13.9658907762998\\
	14	-13.416432576114\\
	16	-12.1728412073277\\
	17	-11.623164049641\\
	18	-11.0112129997509\\
	19	-10.4483957080936\\
	20	-9.7956864362921\\
	21	-9.16189266607056\\
	22	-8.61394787631413\\
	23	-8.17184191802369\\
	24	-7.90800991821546\\
	25	-7.7361548447321\\
	26	-7.61501509854793\\
	27	-7.56602186566748\\
	28	-7.53486915930023\\
	29	-7.46369537051277\\
	30	-7.59070925349975\\
	31	-7.50403502633625\\
	32	-7.5351111139551\\
	33	-7.5455238284306\\
	34	-7.45663606253176\\
	35	-7.39859895081558\\
	36	-7.47486583683151\\
	37	-7.49557886251533\\
	38	-7.34107330970065\\
	39	-7.34897943726864\\
	40	-7.47664235416904\\
	41	-7.63904076563048\\
	42	-7.66187632463232\\
	44	-7.80480458225621\\
	45	-7.81860349262128\\
	46	-7.60857475589485\\
	47	-7.46392427772895\\
	48	-7.34666472333366\\
	49	-7.33454563804689\\
	50	-7.22759215132039\\
	51	-7.2493606621761\\
	52	-7.31419638433091\\
	54	-7.30058224098957\\
	55	-7.31483539709183\\
	56	-7.21643680704773\\
	57	-7.23826744366995\\
	58	-7.3745553260577\\
	59	-7.363496311354\\
	60	-7.25360906249583\\
	61	-7.15502660101168\\
	62	-7.08788657115727\\
	63	-7.01283122133155\\
	64	-6.79402547373103\\
	65	-6.69843747954299\\
	66	-6.73557075303192\\
	67	-6.80513621157112\\
	68	-6.91392813510484\\
	69	-6.92691076951117\\
	70	-6.82851486126607\\
	71	-6.80379733013075\\
	72	-6.8109070259008\\
	73	-6.73392282268988\\
	74	-6.82518485793558\\
	76	-6.91927544967108\\
	77	-6.86466361839773\\
	78	-6.84455341881042\\
	79	-6.90108494780135\\
	80	-6.99699917715071\\
	81	-6.92988132917451\\
	82	-6.99204363950849\\
	83	-6.96730289175804\\
	84	-7.10087551693555\\
	85	-7.02517239420551\\
	86	-7.08948329571849\\
	87	-7.06139959928613\\
	88	-7.00496777863584\\
	89	-7.00356934574066\\
	90	-7.0341420289982\\
	91	-7.03510629082595\\
	92	-7.1246473393886\\
	93	-7.18153983756024\\
	94	-7.09938371573901\\
	95	-7.09798292375727\\
	96	-7.10778790740626\\
	97	-7.03736493410922\\
	98	-7.12967260689331\\
	99	-7.00664249992768\\
	100	-6.92959420497448\\
};

\legend{ICO${}_\mathrm{exp}$, $\mathrm{C}_\mathrm{coarse,long}$ with $a_\mathrm{idle}$, $\mathrm{C}_\mathrm{coarse,long}$ without $a_\mathrm{idle}$}
\end{axis}
\end{tikzpicture}%
    \setlength{\belowcaptionskip}{-10pt}
    \caption{Combined setup $\mathrm{C}_\mathrm{coarse,long}$ in the swing-up phase.}			
    \label{fig:kombi1}	
\end{figure}
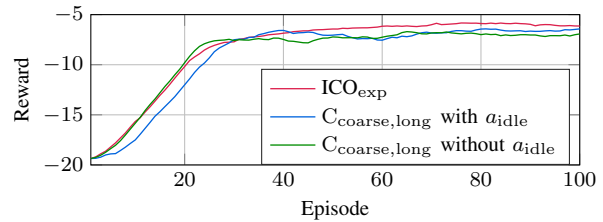
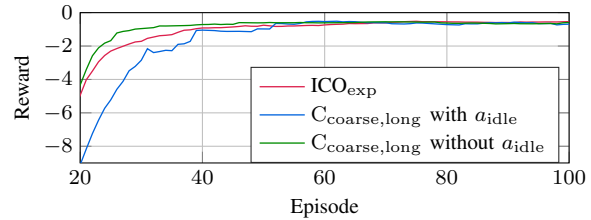
\begin{figure}[h]
    \footnotesize
    \centering 						 
%
%
%
\begin{tikzpicture}

\begin{axis}[%
width=7.1cm,
height=4cm,
scale only axis,
xmin=20,
xmax=100,
xlabel style={font=\color{white!15!black}},
xlabel={Episode},
ymin=-9,
ymax=0,
ylabel style={font=\color{white!15!black}},
ylabel={Reward},
xlabel near ticks,
ylabel near ticks,
axis background/.style={fill=white},
xmajorgrids,
ymajorgrids,
legend style={at={(0.98,0)},anchor=south east,legend cell align=left,align=left,draw=white!15!black},
]
\addplot [color=mumred, mark=square, mark repeat=20, line width=0.5pt]
  table[row sep=crcr]{%
19	-5.81241952544134\\
20	-4.94486393229826\\
21	-4.01857651184031\\
22	-3.4988152484046\\
23	-2.94540150735415\\
24	-2.56713337790274\\
25	-2.30154640240976\\
26	-2.14946046821584\\
28	-1.87848574172729\\
29	-1.756285181762\\
30	-1.71761911598665\\
31	-1.54043273689186\\
33	-1.38185993394298\\
34	-1.35669733403195\\
35	-1.31868379232156\\
36	-1.17789478686821\\
37	-1.05417858753813\\
38	-1.03024240624579\\
39	-0.960547821402457\\
40	-0.916770114129349\\
41	-0.917263352817443\\
42	-0.902348017923131\\
43	-0.90029547748874\\
45	-0.856433106800083\\
46	-0.827643791558145\\
47	-0.7878273632587\\
48	-0.814336911472211\\
49	-0.816568776496965\\
50	-0.752368173719447\\
52	-0.825756556461258\\
53	-0.821811900333074\\
54	-0.800150755039596\\
55	-0.798210391178088\\
57	-0.769275898154064\\
58	-0.762297052090446\\
59	-0.772708324894779\\
60	-0.73530779828323\\
61	-0.71319667162723\\
62	-0.698709005470576\\
63	-0.688569940514839\\
64	-0.653281755463098\\
65	-0.660392616923517\\
66	-0.655372112540434\\
67	-0.652525246780897\\
68	-0.614683807578871\\
69	-0.610784838364552\\
70	-0.614569202040599\\
71	-0.573684538628882\\
72	-0.568305438451631\\
73	-0.541987491244498\\
75	-0.517280593771176\\
76	-0.542914486524168\\
77	-0.541920572302843\\
78	-0.547456440340227\\
79	-0.547123810023749\\
81	-0.556312226102364\\
82	-0.557162725743055\\
83	-0.56861808193338\\
84	-0.567865105675139\\
85	-0.569660618694158\\
86	-0.58084236108806\\
87	-0.583879838080478\\
88	-0.594608334919471\\
89	-0.588591760909154\\
90	-0.59107348994192\\
91	-0.597003882698715\\
92	-0.55857784078664\\
93	-0.569541437483096\\
94	-0.562297931840988\\
95	-0.563854826868322\\
96	-0.557404545082434\\
97	-0.556542569201298\\
98	-0.558781285439295\\
99	-0.545667135069294\\
100	-0.537981911708115\\
};

\addplot [color=mumblue, mark=o, mark repeat=20, line width=0.5pt]
table[row sep=crcr]{%
	1	-19.36\\
	2	-19.3207851988174\\
	3	-19.1674246129114\\
	4	-18.9351475578679\\
	5	-18.5007831538018\\
	6	-18.1936992363179\\
	7	-17.7105156076965\\
	8	-17.3896735343686\\
	9	-16.8670135915953\\
	10	-16.2863265340574\\
	11	-16.0237856195245\\
	12	-15.169660624049\\
	13	-14.2726755655942\\
	14	-13.3978342614476\\
	15	-12.457455879695\\
	16	-11.933233041303\\
	17	-11.1128499090351\\
	18	-10.328281252569\\
	19	-10.0444857080932\\
	20	-9.10598360639017\\
	21	-8.15683248213088\\
	22	-7.24206401078463\\
	23	-6.44025728242652\\
	24	-5.72885082338607\\
	25	-5.22275001668784\\
	26	-4.58888920961746\\
	27	-4.11810266119129\\
	28	-3.48209425769488\\
	29	-3.22088069041193\\
	30	-2.85027292902824\\
	31	-2.15893813106413\\
	32	-2.39213091560001\\
	33	-2.32895419300407\\
	34	-2.25488618853396\\
	35	-2.27841242773452\\
	36	-1.89825357862664\\
	37	-1.86600052572751\\
	38	-1.72039755262836\\
	39	-1.06677422522908\\
	40	-1.04723629924995\\
	42	-1.07465393192327\\
	43	-1.11160095827049\\
	44	-1.12382481758783\\
	46	-1.11940159673527\\
	48	-1.14010227273486\\
	49	-0.973034558393351\\
	51	-0.979186323408015\\
	52	-0.648171471466839\\
	53	-0.655841773097805\\
	54	-0.682185750754684\\
	55	-0.695859605479399\\
	56	-0.647351953107886\\
	57	-0.552288816957784\\
	58	-0.526633272560076\\
	59	-0.520494451618305\\
	60	-0.535730880676311\\
	62	-0.514541585084672\\
	63	-0.551519017334215\\
	64	-0.533859252752322\\
	65	-0.543219502425458\\
	67	-0.6074529018249\\
	68	-0.596756331323206\\
	69	-0.608816731426657\\
	70	-0.608314394144301\\
	71	-0.664534486280431\\
	73	-0.679100319644505\\
	74	-0.66087970528865\\
	75	-0.618420584168376\\
	76	-0.624852047870846\\
	80	-0.717368511617778\\
	82	-0.737705810568599\\
	83	-0.665093154538638\\
	84	-0.672435218864862\\
	85	-0.670855281491001\\
	86	-0.632603024280812\\
	87	-0.605349518104745\\
	88	-0.612522141836521\\
	89	-0.597419405330825\\
	90	-0.618801428227783\\
	91	-0.568851405810122\\
	92	-0.594345859177011\\
	93	-0.607831638885358\\
	94	-0.603994982521399\\
	95	-0.657108253727699\\
	97	-0.64752649016593\\
	98	-0.733011178817435\\
	99	-0.702371486822656\\
	100	-0.690554518675171\\
};
\addplot [color=mumgreen, mark=diamond, mark repeat=20,  line width=0.5pt]
table[row sep=crcr]{%
	1	-19.36\\
	2	-19.2606652211699\\
	3	-18.7873771155248\\
	4	-18.1457753351907\\
	5	-17.3526386952911\\
	6	-16.9777257149186\\
	7	-16.1351951122846\\
	8	-15.2372985346306\\
	9	-14.3873545518237\\
	10	-13.4753928089166\\
	11	-12.5762814719846\\
	12	-11.6590165331887\\
	13	-10.8124116153871\\
	15	-8.93820017580912\\
	16	-8.01507144527901\\
	18	-6.15015618400557\\
	20	-4.3307250587203\\
	21	-3.40443766910806\\
	22	-2.58284263487802\\
	23	-2.10919373726421\\
	24	-1.82622873580902\\
	25	-1.67693845446698\\
	26	-1.21858554715368\\
	27	-1.12480028774439\\
	28	-1.07518108343167\\
	29	-0.985586913246294\\
	32	-0.898247875321886\\
	33	-0.797697479822887\\
	34	-0.799782128870135\\
	37	-0.775515174165776\\
	38	-0.764281921027987\\
	39	-0.731700646500926\\
	40	-0.712694008410963\\
	41	-0.708251294439165\\
	42	-0.686690276822532\\
	43	-0.710819779747482\\
	44	-0.694641463212818\\
	45	-0.700759260896291\\
	46	-0.598638067810782\\
	49	-0.612142948287243\\
	50	-0.608396250450994\\
	51	-0.591175725214242\\
	52	-0.604115479236086\\
	55	-0.603277838989769\\
	56	-0.59061000476099\\
	57	-0.605208770113862\\
	58	-0.604794223615201\\
	59	-0.613549358108003\\
	65	-0.593122967685829\\
	67	-0.558847147064171\\
	70	-0.547238222012339\\
	71	-0.556216957292492\\
	72	-0.549682626087488\\
	76	-0.570435067233234\\
	78	-0.559667272325072\\
	79	-0.61080508230647\\
	81	-0.638178895568757\\
	82	-0.640808917923081\\
	84	-0.622754081245802\\
	85	-0.62740108039732\\
	87	-0.649271601968607\\
	88	-0.673955247238695\\
	90	-0.659475382571884\\
	91	-0.649424213679026\\
	92	-0.66464592084877\\
	94	-0.668041055127262\\
	97	-0.648847339567297\\
	98	-0.66963492857397\\
	99	-0.608114842835747\\
	100	-0.58747907886233\\
};

\legend{ICO${}_\mathrm{exp}$, $\mathrm{C}_\mathrm{coarse,long}$ with $a_\mathrm{idle}$, $\mathrm{C}_\mathrm{coarse,long}$ without $a_\mathrm{idle}$}
\end{axis}
\end{tikzpicture}%
    \setlength{\belowcaptionskip}{-10pt}
    \caption{Combined setup $\mathrm{C}_\mathrm{coarse,long}$ in the hold phase.}			
    \label{fig:kombi2}	
\end{figure}
%
%
The swing-up phase shows higher rewards when including action $a_\mathrm{idle}$. 
However, by the end of the learning period, the controller received less overall reward compared to the ICO${}_\mathrm{exp}$. 
On the other hand, during the hold phase, the reward increase is steeper without $a_\mathrm{idle}$. 
In conclusion, no clear advantage of the new combination with respect to the ICO${}_\mathrm{exp}$ can be seen from either experiment, since no configuration shows best results for both, swing-up and hold phase, cf.~Fig.~\ref{fig:kombi1} and \ref{fig:kombi2} respectively.

\begin{figure}
    \footnotesize
    \centering
    \includegraphics[width=1.05\columnwidth]{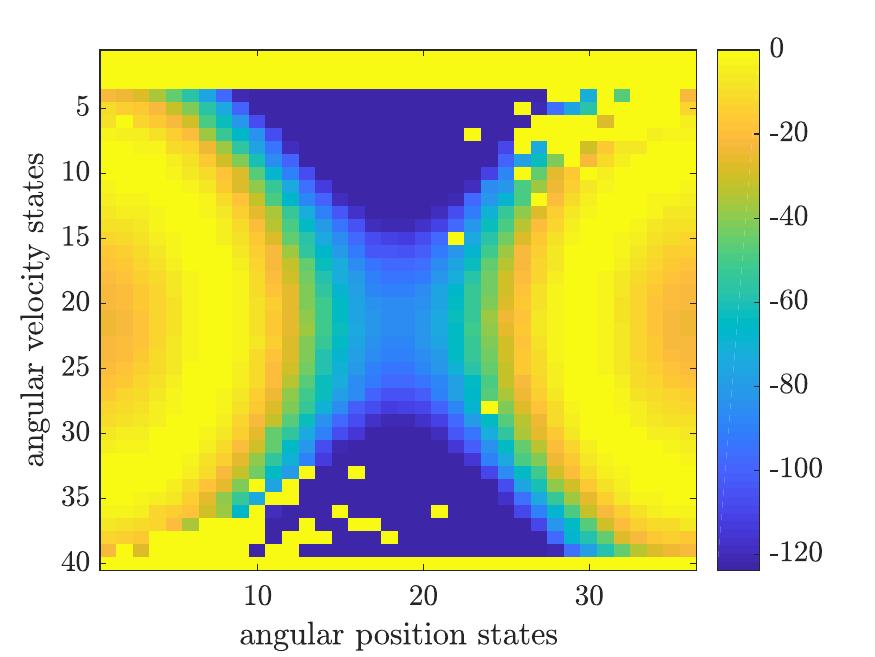}
    \setlength{\belowcaptionskip}{-0pt}
    \caption{State values after 100 episodes.\\$ $}
    \label{fig:state_values_experiment}
\end{figure}

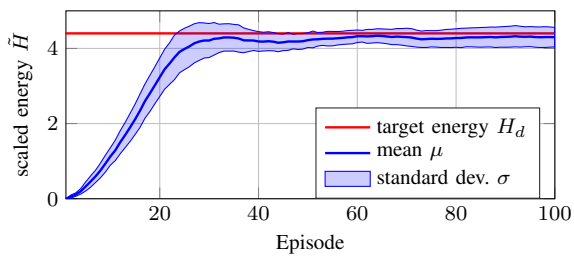
\begin{figure}
    \centering	
    \footnotesize
    \begin{tikzpicture}
\begin{axis}[%
width=7.2cm,
height=2.5cm,
scale only axis,
xmin=1,
xmax=100,
xlabel style={font=\color{white!15!black},font=\footnotesize},
xlabel={Episode},
ymin=0,
ymax=5,
xlabel near ticks,
ylabel near ticks,
ylabel style={font=\color{white!15!black},font=\footnotesize},
ylabel={scaled energy $\tilde{H}$},
axis background/.style={fill=white},
tick label style={font=\footnotesize},  
xmajorgrids,
ymajorgrids,
legend style={at={(0.97,0)},anchor=south east,legend cell align=left,align=left,draw=white!15!black,font=\footnotesize},
]
\addplot [color=red, dashed, line width=0.9pt]
table[row sep=crcr]{%
    0	4.40000000000001\\
    100	4.40000000000001\\
};
\addplot [color=blue, line width=0.9pt]
table[row sep=crcr]{%
    1	0\\
    2	0.0575808021118434\\
    3	0.112204751888285\\
    4	0.197168774911532\\
    5	0.326380748579197\\
    6	0.457807798109087\\
    7	0.595838518694009\\
    8	0.753878522567547\\
    9	0.936239582624111\\
    10	1.12491202871722\\
    11	1.28746056553871\\
    12	1.50420490646658\\
    13	1.69589832300765\\
    14	1.89954977892606\\
    15	2.11244023086978\\
    16	2.34110776249588\\
    17	2.59688599878875\\
    18	2.80223651845819\\
    21	3.44831101361727\\
    22	3.59925562293563\\
    23	3.76436566162586\\
    24	3.90811047604721\\
    25	3.98494652268889\\
    26	4.05278565461715\\
    27	4.13638518508223\\
    28	4.18403706804067\\
    29	4.21074561670419\\
    30	4.21020293222574\\
    31	4.25582902092043\\
    32	4.26673721134314\\
    33	4.29172767948585\\
    34	4.29219938454597\\
    35	4.28455732560994\\
    36	4.26343765814332\\
    37	4.21532542842894\\
    39	4.20932641405767\\
    40	4.1843422175305\\
    41	4.18092388418231\\
    42	4.19164928496023\\
    43	4.17927516259664\\
    44	4.1506086567142\\
    45	4.16279231053552\\
    46	4.18425146086288\\
    48	4.1831729979148\\
    49	4.19120272777005\\
    50	4.22057774590714\\
    51	4.23863180108063\\
    52	4.2429686164682\\
    53	4.255442275468\\
    54	4.25989353785594\\
    55	4.26012938603087\\
    57	4.28035768169018\\
    58	4.29442774186019\\
    59	4.30182492436485\\
    60	4.32403697584952\\
    61	4.32457969523153\\
    62	4.31971225551276\\
    63	4.32500158455443\\
    64	4.33816055608878\\
    65	4.33075727589683\\
    67	4.30418831785796\\
    68	4.31502677334237\\
    69	4.30715672114017\\
    70	4.30160820611643\\
    71	4.28897343009103\\
    72	4.26127735496453\\
    73	4.24302531265695\\
    75	4.25702293873717\\
    76	4.25417133814977\\
    77	4.25561662318164\\
    78	4.26552170342939\\
    79	4.26672601815562\\
    80	4.27680395558713\\
    81	4.27811139660092\\
    82	4.29103218003527\\
    83	4.29553878471512\\
    84	4.2935145504604\\
    85	4.29470486352727\\
    87	4.30408023603451\\
    88	4.29517667816066\\
    89	4.30420442083059\\
    90	4.30721737879068\\
    91	4.31271566704204\\
    92	4.32453647551723\\
    93	4.31258141357507\\
    94	4.30953249754515\\
    95	4.31454161874494\\
    96	4.30320223674633\\
    97	4.29810184437071\\
    98	4.30191406996171\\
    99	4.30234959626061\\
    100	4.29896611527204\\
};

\addplot[area legend, draw=blue, fill=blue, fill opacity=0.2]
table[row sep=crcr] {%
    x	y\\
    1	0\\
    2	0.0976474545809541\\
    3	0.158026265130578\\
    4	0.27015003426183\\
    5	0.448291687057373\\
    6	0.607727908024885\\
    7	0.75886046171355\\
    8	0.950358394904157\\
    9	1.16148711966677\\
    10	1.35731338653984\\
    11	1.5610224074851\\
    12	1.81098442598781\\
    13	2.03231944779823\\
    14	2.27414570393428\\
    15	2.5179997848792\\
    16	2.78623649911676\\
    17	3.05394428900414\\
    18	3.28408346458818\\
    19	3.51101341635645\\
    20	3.7464614522714\\
    21	3.98044324400346\\
    22	4.15308508705253\\
    23	4.35730413157774\\
    24	4.4640776957595\\
    25	4.53614779996909\\
    26	4.59305996824385\\
    27	4.65688766286027\\
    28	4.68621248348611\\
    29	4.67735710159251\\
    30	4.66289151589821\\
    31	4.68909590055101\\
    32	4.64465319619341\\
    33	4.65423510245435\\
    34	4.65711396711384\\
    35	4.63990608425513\\
    36	4.62679354472553\\
    37	4.54534323613703\\
    38	4.50769363670218\\
    39	4.47369118913833\\
    40	4.44703295760152\\
    41	4.4303329140549\\
    42	4.44401177329253\\
    43	4.42247576235048\\
    44	4.39015667455595\\
    45	4.38379845497936\\
    46	4.4202929498151\\
    47	4.37855689108312\\
    48	4.39504739422639\\
    49	4.39984164034115\\
    50	4.41620547053403\\
    51	4.46130546814263\\
    52	4.42326682262001\\
    53	4.43089674952053\\
    54	4.45484232837855\\
    55	4.45176614827007\\
    56	4.45063058255805\\
    57	4.45694053010408\\
    58	4.47382349485322\\
    59	4.481722181886\\
    60	4.49026406409319\\
    61	4.4798672843276\\
    62	4.50548702852105\\
    63	4.50650679472007\\
    64	4.52305961164535\\
    65	4.51988552516944\\
    66	4.51261745816041\\
    67	4.50781306067548\\
    68	4.52913658122373\\
    69	4.52809694394654\\
    70	4.53691726868722\\
    71	4.5255567056307\\
    72	4.48810791219772\\
    73	4.46057468342006\\
    74	4.47606916621269\\
    75	4.47293012029848\\
    76	4.47354534618202\\
    77	4.48679762590829\\
    78	4.50562355254981\\
    79	4.51049672917644\\
    80	4.52315057673917\\
    81	4.53481071669995\\
    82	4.53890611957933\\
    83	4.54757238215383\\
    84	4.55021749138646\\
    85	4.5564707389271\\
    86	4.56062931428229\\
    87	4.56907387730399\\
    88	4.56678199735468\\
    89	4.57572035869222\\
    90	4.58337767118437\\
    91	4.5916924655874\\
    92	4.60741446861122\\
    93	4.58985214976199\\
    94	4.57492580135159\\
    95	4.58096401869961\\
    96	4.57397124981967\\
    97	4.57160800802145\\
    98	4.57855942601149\\
    99	4.57110573171739\\
    100	4.55911156164811\\
    100	4.03882066889598\\
    99	4.03359346080382\\
    98	4.02526871391193\\
    97	4.02459568071998\\
    96	4.03243322367298\\
    95	4.04811921879026\\
    94	4.0441391937387\\
    93	4.03531067738816\\
    92	4.04165848242325\\
    91	4.03373886849669\\
    90	4.031057086397\\
    89	4.03268848296896\\
    88	4.02357135896664\\
    87	4.03908659476501\\
    86	4.03766668485697\\
    85	4.03293898812745\\
    84	4.03681160953435\\
    83	4.04350518727642\\
    82	4.04315824049122\\
    81	4.02141207650189\\
    80	4.03045733443508\\
    79	4.02295530713479\\
    78	4.02541985430896\\
    77	4.024435620455\\
    76	4.03479733011754\\
    75	4.04111575717586\\
    74	4.02422895467468\\
    73	4.02547594189385\\
    72	4.03444679773135\\
    71	4.05239015455137\\
    70	4.06629914354565\\
    69	4.08621649833382\\
    68	4.10091696546102\\
    67	4.10056357504044\\
    66	4.12369385087846\\
    65	4.14162902662422\\
    64	4.15326150053222\\
    63	4.14349637438878\\
    62	4.13393748250444\\
    61	4.16929210613546\\
    60	4.15780988760585\\
    59	4.12192766684369\\
    58	4.11503198886717\\
    57	4.10377483327627\\
    56	4.09096091670978\\
    55	4.06849262379166\\
    54	4.06494474733334\\
    53	4.07998780141548\\
    52	4.06267041031641\\
    51	4.01595813401864\\
    50	4.02495002128024\\
    49	3.98256381519893\\
    48	3.9712986016032\\
    47	3.98748955872971\\
    46	3.94820997191068\\
    45	3.94178616609169\\
    44	3.91106063887245\\
    43	3.93607456284279\\
    42	3.93928679662793\\
    41	3.9315148543097\\
    40	3.92165147745947\\
    39	3.94496163897702\\
    38	3.91612956681985\\
    37	3.88530762072087\\
    36	3.90008177156111\\
    35	3.92920856696474\\
    34	3.92728480197808\\
    33	3.92922025651735\\
    32	3.88882122649286\\
    31	3.82256214128986\\
    30	3.75751434855327\\
    29	3.74413413181587\\
    28	3.68186165259522\\
    27	3.61588270730418\\
    26	3.51251134099046\\
    25	3.43374524540868\\
    24	3.35214325633493\\
    23	3.17142719167398\\
    22	3.04542615881873\\
    21	2.91617878323109\\
    20	2.72091066423256\\
    19	2.52434923353555\\
    18	2.3203895723282\\
    17	2.13982770857334\\
    16	1.89597902587501\\
    15	1.70688067686034\\
    14	1.52495385391785\\
    13	1.35947719821709\\
    12	1.19742538694537\\
    11	1.01389872359232\\
    10	0.892510670894605\\
    9	0.710992045581465\\
    8	0.557398650230949\\
    7	0.432816575674477\\
    6	0.307887688193297\\
    5	0.204469810101021\\
    4	0.124187515561229\\
    3	0.06638323864598\\
    2	0.0175141496427335\\
    1	0\\
}--cycle;
\legend{target energy ${H}_d $, mean $\mu$ , standard dev. $\sigma$}
\end{axis}
\end{tikzpicture}%
    
    \setlength{\belowcaptionskip}{-10pt}
    \caption{Scaled energy $\tilde{H}$ over 10 runs with $c_\mathrm{exp}=7.06$.}%
    \label{fig:sumEnergy_simulation}%
\end{figure}

As can be seen in Fig.~\ref{fig:state_values_experiment}, the corresponding  value function qualitatively resembles the simulation results from Fig.~\ref{fig:valueFunction}, which validates the correct measurement of the angular position $\theta$ and velocity $\dot{\theta}$ of the first pendulum and the corresponding energy as given in Eq.\,\eqref{eq:hamiltonian}. 
The scattered yellow dots in Fig.~\ref{fig:state_values_experiment} represent states which were not visited up to episode 100. Thus, these states still posses their zero value from the initialization step.
It turns out that the RL controller adapts well to different setups and achieves convergence to the desired control in all studied cases.
In Fig.~\ref{fig:k_energy} and \ref{fig:k_phase}, the energy curve and the phase space trajectory of the first pendulum for the 100th episode are illustrated for the case $\mathrm{C}_\mathrm{coarse,long}$ without $ a_\mathrm{idle}$.
\begin{figure}[h]
    \footnotesize
    \begin{minipage}{0.5\textwidth}
        \centering 						 
        \input{kombi_energy.tikz}
        \caption{Energy of the first pendulum in the 100th episode.}			
        \label{fig:k_energy}
    \end{minipage}
    \begin{minipage}{0.5\textwidth}
        \centering 						 
        \input{kombi_phase.tikz}
        \setlength{\belowcaptionskip}{-0pt}
        \caption{Trajectory of the first pendulum in the 100th episode.}			
        \label{fig:k_phase}
    \end{minipage}
\end{figure}
%
%
Moreover, the average of ten corresponding experimental runs shows that the mean energy converges quickly within the first 30 episode, see Fig.~\ref{fig:sumEnergy_simulation}.

\subsection{Rotation Control}\label{sec:rotcontrol}

In addition to energy control, further control goals can be considered by designing appropriate reward functions. 
Previously, we chose the energy level in a way such that the first pendulum does not reach the unstable equilibrium position at $ \theta = 180^{\circ} $. 
In the following, we aim to drive the first pendulum towards a desired angular velocity during rotation. 
One possible approach is to enforce the rotation by setting a sufficiently large desired angular velocity. 
For this case $ \widetilde{H} > 2gc_\mathrm{exp}$ has to be approximately fulfilled~\cite{dostal2018pendulum}.  
In the phase portrait, this corresponds to the area outside the separatrix, which is defined by $ \widetilde{H} = 2gc_\mathrm{exp} $, so that the angular velocity $ \dot{\theta} $ does not change its sign and the $ \theta $ axis is not crossed in the phase space. 
The scaled Hamilton function is only used at this point to approximate the required angular velocity.
From Eq.\,\eqref{eq:scaledhamiltonian} we get for the border case $ \widetilde{H} = 2gc_\mathrm{exp} $ which separates the pendulum oscillation from the pendulum rotation
\begin{equation}
\dot{\theta} = \sqrt{4gc_\mathrm{exp}\left(1 - \sin^2\left(\frac{\theta}{2}\right)\right)}.
\end{equation} 
Thus, in this case the angular velocity has a maximum of $ 2 \sqrt{gc_\mathrm{exp}} $ and a minimum of zero. 
Note that the factor $ c_\mathrm{exp} $ is not exactly known. Hence, we define $ \dot{\theta}_d = 3 \sqrt{gc_\mathrm{exp}}$ as simplification. 
In addition, we reformulate the reward function as
\begin{equation}\label{vel_rf}
R(\dot{\theta}) = -(\dot{\theta}_d - |\dot{\theta}|)^2.
\end{equation}
Moreover, the state space domain for the RL controller has to be increased by doubling the width of the considered angular velocity interval to $ [-10,10] \unit{rad/s} $. 
The higher the desired angular velocity $ \dot{\theta}_d $ is set, the more time is required for swing-up phase part of the episode. 
Therefore, we extend the episode length to $ \unit[200]{s} $ and increase the mass $ m_2 $ for this experiment. 
Given the results from Sec.\,\ref{sec:energycontrol}, we again double the step size to $ \Delta\theta=20^\circ$. 
Figure~\ref{fig:rot_reward} depicts the average reward received in each episode after $ \unit[20]{s} $. 
\begin{figure}
    \footnotesize
    \begin{minipage}{0.5\textwidth}
        \centering 						 
%
%
%
\begin{tikzpicture}

\begin{axis}[%
width=7.2cm,
height=2.5cm,
scale only axis,
xmin=1,
xmax=50,
xlabel style={font=\color{white!15!black}},
xlabel ={Episode},
ymin=-70,
ymax=-20,
xlabel near ticks,
ylabel near ticks,
ylabel style={font=\color{white!15!black}},
ylabel={Reward},
axis background/.style={fill=white},
xmajorgrids,
ymajorgrids,
]
\addplot [color=mumblue]
  table[row sep=crcr]{%
1	-63.5687999999976\\
2	-53.2843757759311\\
3	-49.1155770462331\\
4	-48.9356957258882\\
5	-50.592806967083\\
6	-44.3524118917383\\
7	-47.5130823646427\\
8	-44.5858771898798\\
10	-37.5713308043583\\
11	-38.6699855750809\\
12	-40.3085962911379\\
13	-38.2776572773717\\
14	-39.6993297238866\\
15	-40.4452442292576\\
16	-42.2948741045978\\
17	-34.770599921797\\
18	-32.5206807713323\\
19	-32.9347869104003\\
20	-42.0678046872519\\
21	-39.0520350622704\\
22	-38.6899363477143\\
23	-42.9833141455487\\
24	-33.2486245818205\\
25	-37.6079434335622\\
26	-39.5505594391916\\
27	-36.0008832562283\\
28	-32.1845280763029\\
29	-32.0740407535584\\
30	-30.5412749885799\\
31	-30.2629894994192\\
32	-30.6096609759199\\
33	-29.8634566365065\\
34	-35.8695591178939\\
35	-32.5268439036141\\
36	-30.3202687640927\\
37	-30.1168595455862\\
38	-30.9022998743015\\
39	-25.936675697998\\
40	-26.6065239072673\\
41	-27.1431483390369\\
42	-29.5186184829701\\
43	-25.8812629215952\\
44	-27.5908697766079\\
45	-26.9439112531229\\
46	-27.9912112015303\\
47	-30.4174060254175\\
48	-27.6066371503897\\
49	-24.217153037714\\
50	-24.9560067917425\\
};

\end{axis}
\end{tikzpicture}%
        \setlength{\belowcaptionskip}{-0pt}
        \caption{Mean reward received after $ \unit[20]{s} $ in each episode.}			
        \label{fig:rot_reward}
    \end{minipage}
    \begin{minipage}{0.5\textwidth}
        \centering 						 
        \input{rot_phase.tikz}
        \setlength{\belowcaptionskip}{-10pt}
        \caption{Phase portrait for rotation control of the pendulum, as obtained in the experiment.}
        \label{fig:rot_phase}
    \end{minipage}
\end{figure}
The learning curve shows a clear upward trend. 
In order to accelerate the swing-up even more, the used action space $ \mathcal{A}_\mathrm{extended} $, see Tab.~\ref{tab:cases}, is expected to improve the result even further.
However, in this case the time step size would have to be at least tripled. 
This would have negative consequences for the determination of the exact state in the RL algorithm. 
In order to achieve higher angular velocities, longer episodes can be selected and the total learning time can be increased even further.
Nevertheless the RL controller manages to bring the first pendulum into rotation with this configuration, as can be seen in the phase portrait depicted in Fig.~\ref{fig:rot_phase} for the 50th episode.

%
The transitions between pendulum motion and rotational motion do not seem to cause any problems for the examined RL algorithm. 
Moreover, the state values are interpreted meaningfully in both operating ranges by the learning algorithm. 
This would be different if a classical control algorithm, such as for example sliding mode control, would be used, in which case the state space regions of different dynamical behavior would have to be known beforehand. 
This is not necessary if the presented RL control is used.

\section{Comparison with sliding mode control}\label{sec:sliding_mode_control}
\textcolor{black}{For the evaluation of the performance of the above RL controller for the Acrobot, we compare the corresponding control results to results of a sliding mode controller.
A sliding mode controller is capable to control nonlinear dynamical systems towards the so called sliding surface.}

\begin{figure}
\footnotesize
	\centering 						 
	\begin{subfigure}[b]{0.9\columnwidth}
     \centering  
     \includegraphics[width=1.07 \columnwidth]{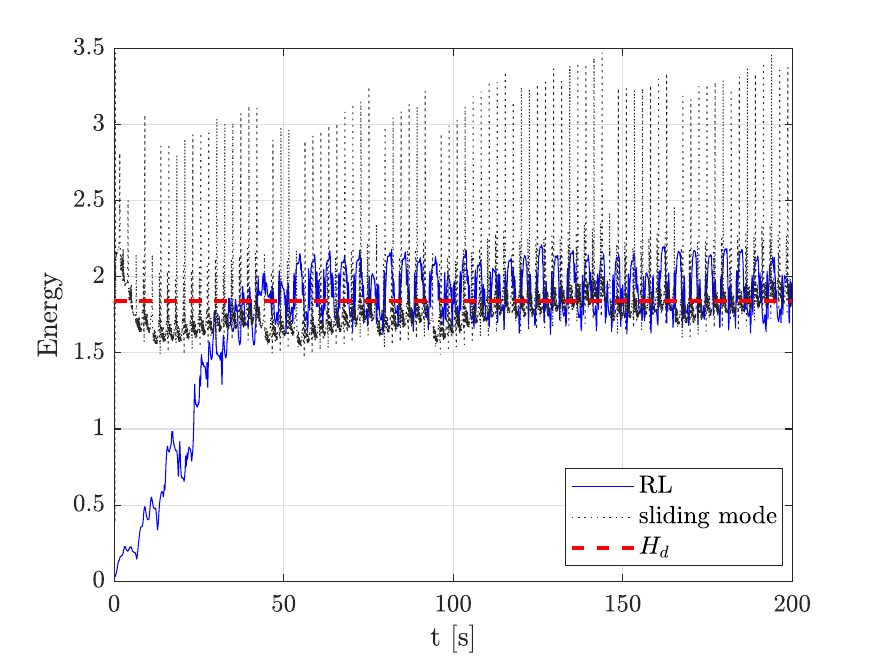}
     \caption{Desired energy level $H_d=0.75\, \alpha$.}
     \label{fig:H_d0.75alpha}
    \end{subfigure}
    \begin{subfigure}[b]{0.9\columnwidth}
         \centering
         \includegraphics[width=1.07 \columnwidth]{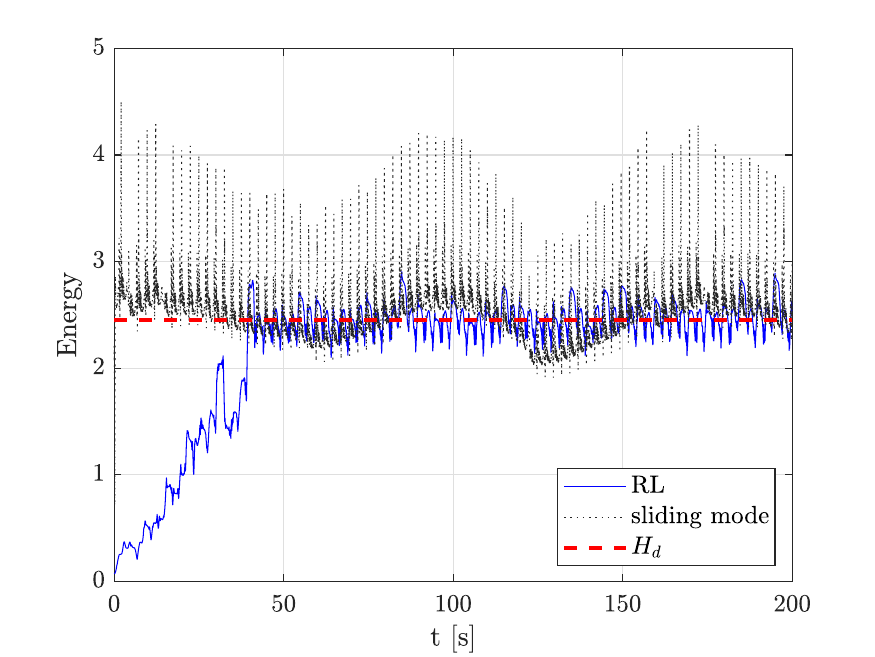}
         \caption{Desired energy level $H_d=1.00 \,\alpha$.}
         \label{fig:H_d1.00alpha}
    \end{subfigure}
    \begin{subfigure}[b]{0.9\columnwidth}
         \centering
         \includegraphics[width=1.07 \columnwidth]{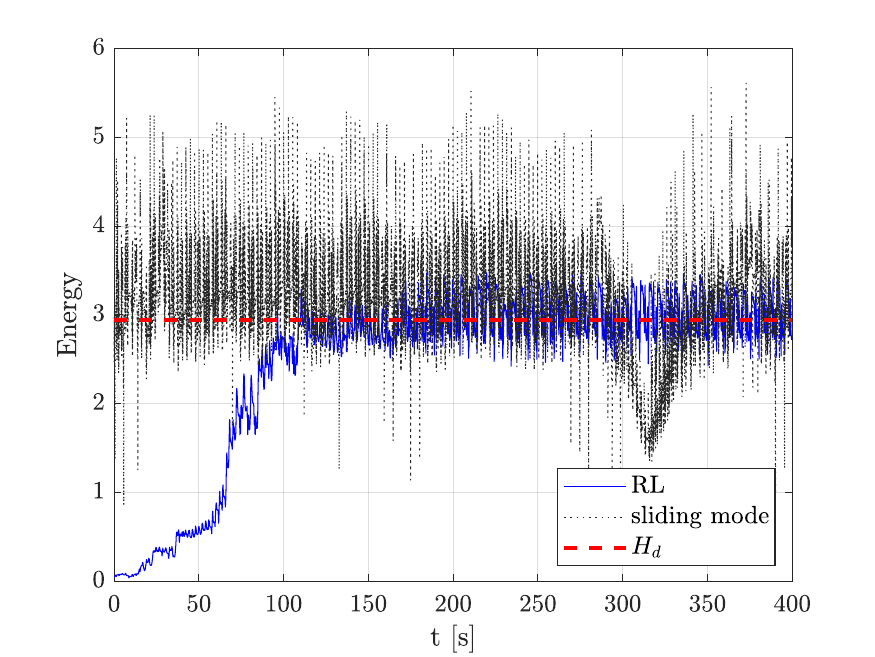}
         	\caption{Desired energy level $H_d=1.20\, \alpha$.}
         	\label{fig:H_d1.20alpha}
    \end{subfigure}
	\caption{Energy time series of the first pendulum for different desired energy levels $H_d$, using the RL controller and the sliding mode controller.}	
	\label{fig:CompareEnergy}	
\end{figure}

\textcolor{black}{Using the theory described in} \cite{slotine1991applied,otto2018real}
\textcolor{black}{, we have developed a sliding mode controller for the Acrobot. For this controller it is necessary to determine a target trajectory $ [\theta_d, \dot{\theta}_d] $ for $ H_d = const. $, which can be obtained using the Hamiltonian~\eqref{eq:hamiltonian}. As shown in}~\cite{dostal:2017b}\textcolor{black}{, for a given energy level $H$ this trajectory is given by
\begin{equation}\label{eq:energytrajectory}
\begin{aligned}
\theta_d(t)=& 2\arcsin(k\mathrm{sn}(\sqrt{\alpha}\,t,k))\\
\dot{\theta}_d(t)=&2k\sqrt{\alpha}\mathrm{cn}(\sqrt{\alpha}\,t,k).
\end{aligned}
\end{equation}
Thereby sn and cn are Jacobi elliptic functions \cite{byrd2013handbook}, $\alpha=g/l_1$ is the eigenfrequency of the first pendulum, and $k=\sqrt{\frac{H}{2\,\alpha}}$ is the elliptic modulus.
With this, the control goal is that the first unactuated pendulum of the Acrobot follows this desired trajectory.
For the comparison of the Acrobot control results, we have used numerical simulations of the RL controller and the sliding mode controller. Thereby the values of the ICO from Section~\ref{sec:sim} have been used.
%
The Fig.~\ref{fig:CompareEnergy} shows the development of the energy $ H $ for different desired energy levels $H_d$. With the eigenfrequency of the first pendulum given by $\alpha=g/l_1$,  the results for $ H_d = 0{.}75\alpha$ are shown in panel (a) of this figure. Since the sliding mode controller is not able to control the swing up starting at low initial energy $H_0$, we have used the initial energy $H_0=0.7\,\alpha$, which is close to the desired energy $H_d$. 
On the other hand, the RL controller also automatically learns the swing up of the first pendulum, which is a benefit in comparison with the sliding mode controller. In order to show this, we have set the initial energy $H_0$ close to zero for the case where the Acrobot is controlled by the RL controller. The results of the RL controller and the sliding mode controller show that they are both able to maintain the energy of the first pendulum at the desired energy level. However, it can also be observed, that the sliding mode controller causes higher energy peaks during the control. In panels (b) and (c) of Fig.~\ref{fig:CompareEnergy} results for higher desired energy levels $ H_d = 1{.}00\alpha$ and $ H_d = 1{.}2\alpha$ are shown, respectively. Higher energy levels lead to higher actuation effort of the second actuated pendulum of the Acrobot, but both controllers are still able to control the energy of the first pendulum at the desired energy $H_d$. In comparison it can be seen that the control effort of the sliding mode controller is higher due to the energy peaks caused during control of the Acrobot.} 

%

\begin{figure}
\footnotesize
	\centering 						 
	\begin{subfigure}[b]{1.02\columnwidth}
     \centering
     \includegraphics[width=1.07 \columnwidth]{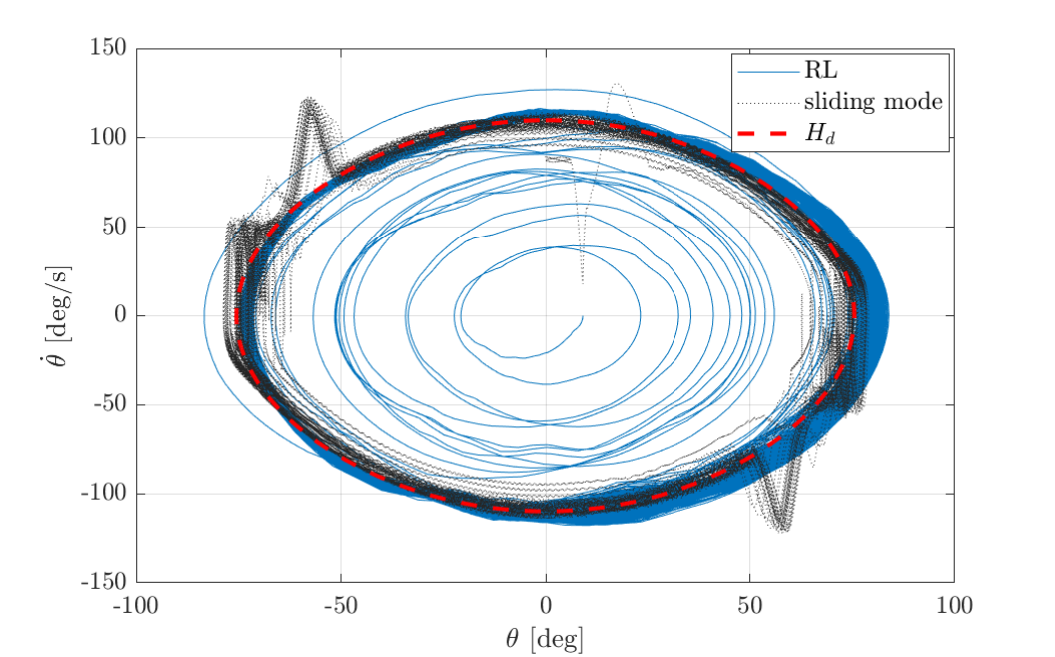}
     \caption{Desired energy level $H_d=0.75\, \alpha$.}
     \label{fig:H_d0.75alpha}
    \end{subfigure}
    \begin{subfigure}[b]{1\columnwidth}
         \centering
         \includegraphics[width=1.07 \columnwidth]{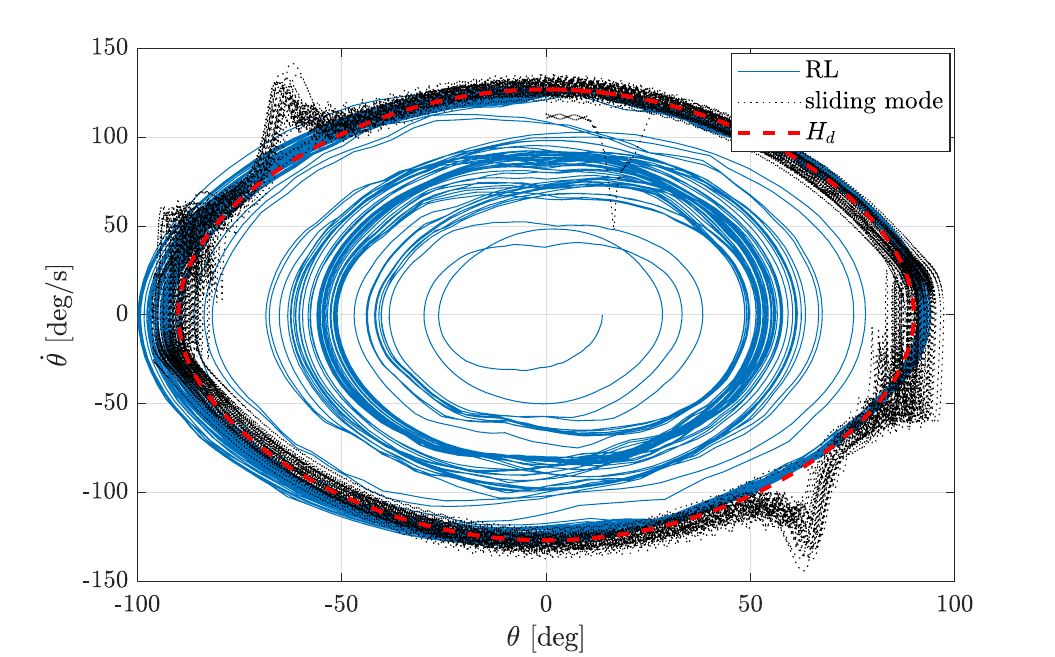}
         \caption{Desired energy level $H_d=1.00\, \alpha$.}
         \label{fig:H_d1.00alpha}
    \end{subfigure}
    \begin{subfigure}[b]{1\columnwidth}
         \centering
         \includegraphics[width=1.07 \columnwidth]{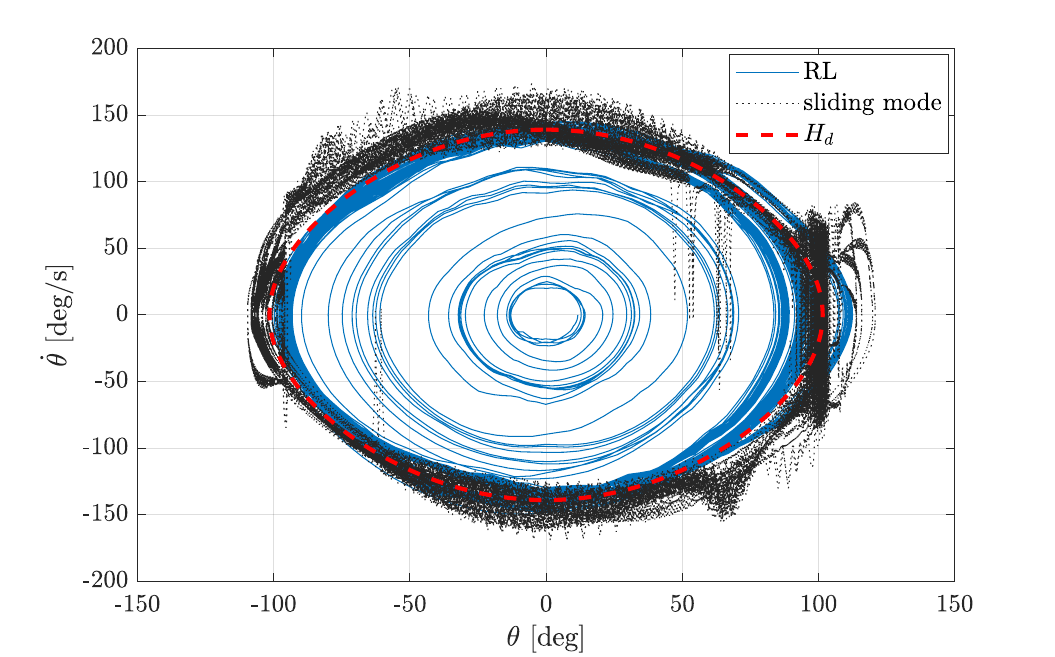}
         	\caption{Desired energy level $H_d=1.20\, \alpha$.}
         	\label{fig:H_d1.20alpha}
    \end{subfigure}
	\caption{Phase portraits of the first not actuated pendulum of the Acrobot for different desired energy levels $H_d$, whereby the results of the RL controller and the sliding mode controller are shown in each panel.}	
	\label{fig:CompareStatespace}	
\end{figure}

\begin{figure}
\footnotesize
	\centering 						 
	\begin{subfigure}[b]{1.02\columnwidth}
     \centering
     \includegraphics[width=1.07 \columnwidth]{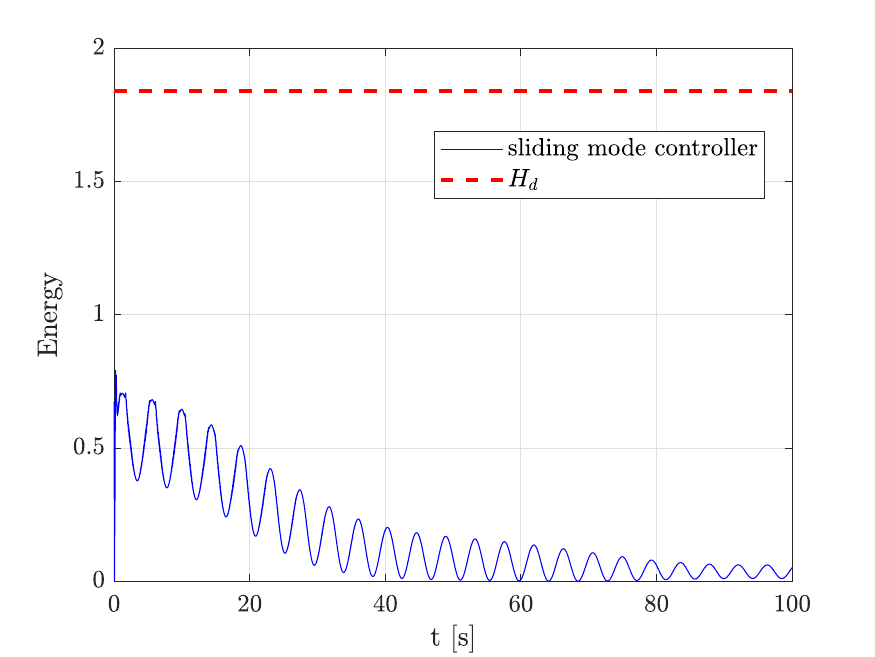}
     \caption{Results of the sliding mode controller for the energy starting at low initial enrgy $H_0=0.25\, \alpha$.}
     \label{fig:H_d0.75alpha}
    \end{subfigure}
    \begin{subfigure}[b]{1\columnwidth}
         \centering
         \includegraphics[width=1.07 \columnwidth]{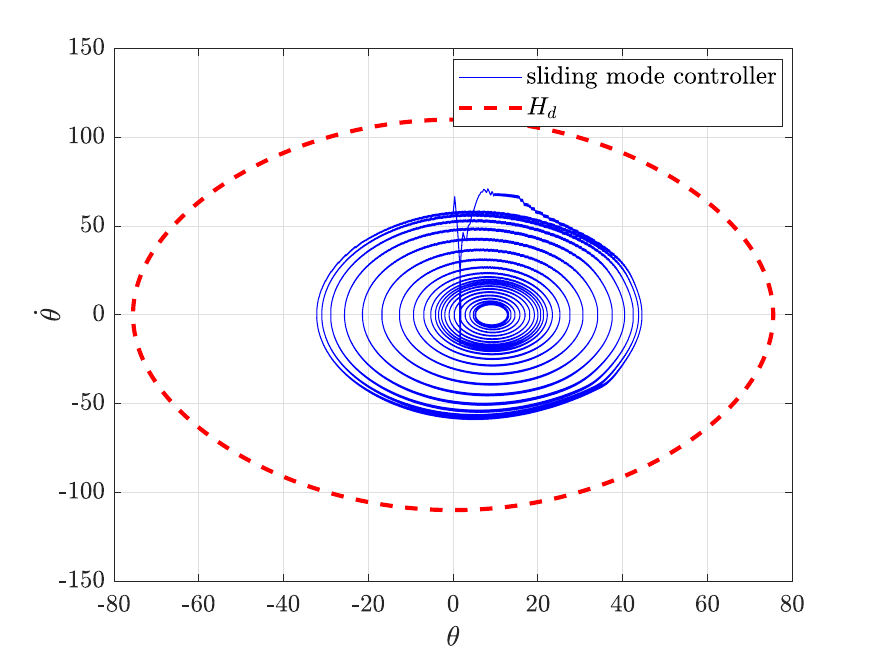}
         \caption{Results of the sliding mode controller for the phase space trajectory starting at low initial enrgy $H_0=0.25\, \alpha$.}
         \label{fig:H_d1.00alpha}
    \end{subfigure}
	\caption{Results of the sliding mode controller for the energy, panel (a), and phase space trajectory, panel (b), of the first not actuated pendulum of the Acrobot for the desired energy level $H_d=0.75\, \alpha$ starting at initial energy $H_0=0.25\, \alpha$.}
	\label{fig:slidingModeFail}	
\end{figure}

\textcolor{black}{The entire simulation result of the Acrobot control can also be illustrated well in a phase portrait. The red dashed curve in panels (a), (b), and (c) of Fig.~\ref{fig:CompareStatespace} corresponds to the trajectory of the respective constant energy levels according to~\eqref{eq:energytrajectory}, such that this curve is the target trajectory. Moreover, the trajectory of the first pendulum controlled by the RL controller is shown in blue. With a sufficiently large starting value of the energy $ H_0 $ the sliding mode controller also manages to steer the pendulum to the required trajectory. The corresponding black dotted trajectories are also shown in panels´(a), (b), and (c) of Fig.~\ref{fig:CompareStatespace} for different desired energy levels $H_d$. If the initial energy $H_0$ of the first pendulum is not close to the desired energy $H_d$, the sliding mode controller fails to reach the desired energy $H_d$, such a case is illustrated in Fig.~\ref{fig:slidingModeFail} for the desired energy $H_d=0.75\, \alpha$ and initial energy $H_0=0.25\, \alpha$ 
%
The desired energy levels $H_d= 0.75 \, \alpha$ and $H_d= 1.0 \, \alpha$ are maintained well by both controllers, as can be observed in panels (a) and (b) of Fig.~\ref{fig:CompareStatespace}.
For the higher energy level $H_d= 1.2 \, \alpha$ the RL controller maintains the desired energy level well, whereas the sliding mode controller encounters more difficulties for maintaining the desired energy level, see panel (c) of Fig.~\ref{fig:CompareStatespace}.}
%
%

\textcolor{black}{In summary, the RL controller, as wel as the sliding mode controller are capeable to control the first pendulum of the Acrobot towards a desired energy level. The RL controller shows even a slightly better performance, since the resulting trajectories are closer to the desired energy level for all tested cases. Moreover, no additional swing up controller has to be designed if the RL controller is used, since the swing up of the first pendulum is learned as well. This is not the case for the sliding mode controller, where an additional controller has to be designed for the swing up of the first pendulum.}

\vspace{0 cm}
\section{CONCLUSION}\label{sec:conclusions}
\textcolor{black}{Extensive results on RL control of an Acrobot in simulation and experiment are obtained in this study.}
Thereby, the Acrobot is stabilized at a desired energy level, which is given by the Hamiltonian of the first unactuated pendulum. \textcolor{black}{Moreover, control of the angular velocity of the first pendulum is achieved as well. 
It is remarkable that the RL controller is even able to stabilize the energy level close to the transition region between libration and rotation of the first pendulum, which is difficult due to the distinct nonlinear dynamical behaviour of the pendulum in these regions.}

\textcolor{black}{Although classical controllers, such as the sliding mode controller, can achieve a comparable performance as the RL controller, we have shown that there are crucial benefits if an RL controller is used. These benefits include that a knowledge of the underlined nonlinear dynamics is not necessary. Moreover, the RL controller learns to control the Acrobot in different phase space regions, such as the phase space regions of rotation and libration of the first unactuated pendulum of the Acrobot. It is also not necessary to develop different RL controllers for these distinct dynamics, as is the case in classical control design, where for example a swing up controller has to be developed.}

The presented experimental results demonstrate the feasibility to deploy the proposed RL algorithm on an Acrobot experimental setup using embedded architecture such as a RaspberryPi.







\bibliographystyle{IEEEtran}
\bibliography{access}

\end{document}